\pdfoutput=1
\documentclass[11pt,a4paper]{article}
\usepackage[hyperref]{emnlp2021}
\usepackage{times}
\usepackage{latexsym}

\usepackage{microtype}

\usepackage[utf8]{inputenc}
\usepackage[T1]{fontenc}
\usepackage{booktabs}
\usepackage{footnote}

\usepackage{amsmath}
\usepackage{amssymb}
\usepackage{comment}
\usepackage{graphicx}
\usepackage{caption, subcaption}
\usepackage{xcolor}

\usepackage{tikz}
\usepackage{xspace}
\usetikzlibrary{positioning}

\hypersetup{
   pdftitle = {A Conditional Generative Matching Model for Multi-lingual Reply Suggestion},
   pdfauthor = {Budhaditya Deb, Guoqing Zheng, Milad Shokouhi, Ahmed Hassan Awadallah}
}

\newcommand{\m}{{\mbox{CGM}}\xspace} 
\newcommand{\mm}{{\mbox{CGM-M}}\xspace}

\title{A Conditional Generative Matching Model for\\Multi-lingual Reply Suggestion}

\author{
Budhaditya Deb$^\dag$ \quad
Guoqing Zheng$^\ddag$ \qquad
Milad Shokouhi$^\dag$ \quad
Ahmed Hassan Awadallah$^\ddag$ \qquad \\
 \quad$^\dag$Microsoft AI \qquad$^\ddag$Microsoft Research\\ 
  \texttt{\{budeb, zheng, milads, hassanam\}@microsoft.com} \\}
\date{}

\begin{document}
    \maketitle
    
\begin{abstract}
  We study the problem of multilingual automated reply suggestions (RS) model serving many languages simultaneously. Multilingual  models are often challenged by model capacity and severe data distribution skew across languages. While prior works largely focus on monolingual models, we propose Conditional Generative Matching models (CGM), optimized within a Variational Autoencoder framework to address challenges arising from multi-lingual RS. CGM does so with expressive message conditional priors, mixture densities to enhance multi-lingual data representation, latent alignment for language discrimination, and effective variational optimization techniques for training multi-lingual RS. The enhancements result in performance that exceed competitive baselines in relevance (ROUGE score) by  more than 10\% on average, and 16\% for low resource languages. CGM also shows remarkable improvements in diversity (80\%) illustrating its expressiveness in representation of  multi-lingual data.
  
\end{abstract}

\section{Introduction}
    Automated reply suggestion (RS) helps users quickly process Email and chats, in popular applications like Gmail, Outlook, Microsoft Teams, and Facebook Messenger, by selecting a relevant reply generated by the system, without having to type in the response. Most existing RS systems are English mono-lingual models \cite{Kannan2016, henderson2017efficient, Deb2019DiversifyingRS, shang2015neural}. We study the problem of creating multilingual RS models serving many languages simultaneously. Compared to mono-lingual models, a universal multilingual model offers several interesting research questions and practical advantages.
    
    Universal models can save compute resources and maintenance overhead for commercial systems supporting  many regions. In addition it can benefit languages with insufficient data by information sharing from high resource languages and thus enhance experiences for users especially in low-language resource regions. We investigate if a single multilingual RS model can replace multiple mono-lingual models with better performance, while overcoming the challenges in model capacity, data skew, and training complexities.
    
    Trivially extending existing mono-lingual RS models to the multilingual setting (e.g. by jointly training with pre-trained multi-lingual encoders) tends to be sub-optimal, as multilingual models suffer from capacity dilution issue \cite{lample2019cross}, where it improves performance on low resource languages while hurting the high resource ones. This arises, not only due to the severe data imbalance and distribution skew across languages, but also due to insufficient capacity and lack of inductive biases in models to represent the multi-modal distribution of languages.  We postulate that deep generative latent variable models with variational auto-encoders (VAE)  \cite{Kingma2013AutoEncodingVB} are better suited to model the complex distribution of multi-lingual data, and be more data efficient for low resource languages. 
    
\begin{figure*}[tp]
  \begin{subfigure}[t]{0.3\textwidth}
    \centering
    \begin {tikzpicture}[-latex, auto, on grid , semithick , var/.style ={circle , draw, text=black , minimum width =0.6 cm, inner sep=0pt}]
      \node[var] (z) {$z$};
      \node[var,fill=gray] (x) [right=of z] {$\Theta_M$};
      \node[var] (y) [right=of x] {$\Theta_{R}$};
      \path (z) edge (x);
      \path (x) edge (y);
      \path (z) edge [bend left=45] (y);
    \end{tikzpicture}
    \caption{Generative model for MCVAE}
    \label{fig:mcvae_model}
  \end{subfigure}
  ~
  \begin{subfigure}[t]{0.3\textwidth}
    \centering
    \begin{tikzpicture}[-latex, auto, on grid , semithick , every node, var/.style ={circle, draw, text=black, minimum width =0.6 cm, inner sep=0pt}]
      \node[var,fill=gray] (x) {$\Theta_M$};
      \node[var] (z) [right=of x] {$z$};
      \node[var] (y) [right=of z] {$\Theta_{R}$};
      \path (x) edge (z);
      \path (z) edge (y);
      \path (x) edge [bend left=45] (y);
    \end{tikzpicture}
    \caption{Generative model for \m}
    \label{fig:cgm_model}
  \end{subfigure}
  ~
  \begin{subfigure}[t]{0.3\textwidth}
    \centering
    \begin{tikzpicture}[-latex, auto, on grid , semithick , every node, var/.style ={circle , draw, text=black, minimum width =0.6 cm, inner sep=0pt}]
      \node[var,fill=gray] (x) {$\Theta_M$};
      \node[var] (c) [right=of x] {$c$};
      \node[var] (z) [right=of c] {$z_c$};
      \node[var] (y) [right=of z] {$\Theta_{R}$};
      \path (x) edge (c);
      \path (x) edge [bend left=30] (z);
      \path (c) edge (z);
      \path (z) edge (y);
      \path (x) edge [bend left=45] (y);
    \end{tikzpicture}
    \caption{Generative model for \mm}
    \label{fig:cgm_gmm_model}
  \end{subfigure}
  
    \label{fig:model1}
    \caption{RS generative models in the continuous space. Text M-R pairs (in discrete space) are encoded into a common continuous space ($\Theta_{M} \sim \Theta_{R}$), where the encoders outputting $\Theta_{M}, \Theta_{R}$ are considered extraneous to the generative model. The generative process is in the continuous space, with $\Theta_{R}$ generated conditioned on the input $\Theta_{M}$ and a Gaussian prior $z$. The figures show three variations of this generative process. In prior work MCVAE, $z$ is sampled independently, while in CGM, it is conditional on $\Theta_M$. CGM-M extends the message conditional prior with a Gaussian Mixture prior $z_c$ and a categorical prior $c$.}\label{fig:all_models}
\end{figure*}
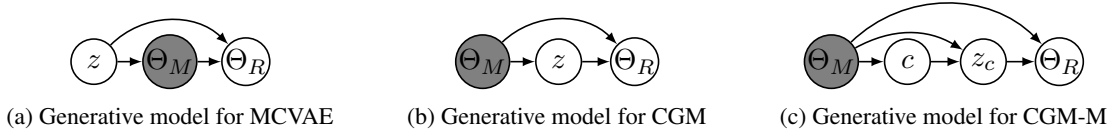
    
 
  To this end, we propose the Conditional Generative Matching Model (CGM), a VAE based retrieval architecture for RS to solve the above challenges. \m enhances multilingual representation through: 1) expressive message conditional priors, 2) multi-component mixture density to represent different modalities of languages, and 3) alignment of latent components for language discrimination. In addition CGM incorporates training optimizations in the form of 1) loss regularizer, 2) learnable weights for loss components, 3) multi-sample loss estimation with variance scaling, and 4) focal loss, all of which lead to balanced representation and smooth convergence, a key challenge for variational training in multilingual settings.  

    We conducted extensive ablation studies and comparisons with two competitive baselines to show the impact of the above optimizations. Universal CGM models improve the relevance of RS (up to 13\% excluding English) with even higher gains coming for low resource languages (16\%), and when using CGM in a monolingual setting (19\%). \m also dramatically increases the diversity of suggested replies by 80\% which is more illustrative of the improved representational capability of CGM in the multi-lingual landscape. \m achieves this with relatively small increase in model sizes compared to the large pre-trained transformer stacks on which it is built, showing the modeling efficiencies that can be achieved through efficient training of latent variable models in a multi-lingual setting.

    \section{Background and Preliminaries} \label{sec:background}
     While RS has been modeled as a sequence to sequence model \cite{Kannan2016}, it more commonly appears as an information retrieval (IR) system by ranking responses from a fixed set \cite{henderson2017efficient,Henderson2019b, QiYingSRNAACL2021, ProdChatbotRetrieval, zhou-etal-2016-multi, zhou-etal-2018-multi} due to better control over quality and relevance for practical systems. We briefly describe two retrieval architectures from prior literature which serves as the baselines for our multilingual RS model.  

    \textbf{Matching model} \cite{henderson2017efficient,QiYingSRNAACL2021} consists of two parallel encoders $[f_{\varphi_M}, f_{\varphi_R}]$ to encode message and reply (M-R) pairs into a common encoding space, $[\Theta_M, \Theta_R]$ and trained to maximize a normalized dot product $D = \Theta_M^\top\Theta_{R}$ between the M-R encodings. During prediction, the model finds the nearest neighbors of $\Theta_M$ with precomputed encodings from a fixed response set $R_{[s]}$. A language model bias is typically added to promote more common responses.  The matching architecture is summarized as: 
    
    {\small
        \begin{gather}
             \mathcal{L}(\Theta_R|\Theta_{M})=\log\frac{e^{D(\Theta_M,\Theta_{R})}} {\sum_{r\in R_{[s]}}e^{D(\Theta_M,\Theta_{r})}} \label{eq:match_loss}\\
            \text{Prediction}: Top_k \{\Theta^\top_M \Theta_r + \alpha LM(r)|r\in R_{[s]}\} \label{eq:match_score}
        \end{gather}
    }
    \textbf{Matching Conditional VAE (MCVAE)} \cite{Deb2019DiversifyingRS} induces a deep generative latent variable model on the matching architecture, where a candidate response encoding is generated with $\Theta_{R'}=g_w(\Theta_M, z)$ conditioned on a latent prior $z \sim \mathcal{N}(0, I)$. The generated $\Theta_{R'}$ is used to match an actual response vector $\Theta_{R}$ from the fixed response set. The generative model of MCVAE is shown in figure \ref{fig:mcvae_model}. In MCVAE, the encoders $[f_{\varphi_M}, f_{\varphi_R}]$ are pretrained using the matching formulation and kept frozen during the training. For prediction, MCVAE samples response vectors from $g_{w}$ followed by scoring (eq \ref{eq:match_score}) and a voting technique to rank replies over a fixed response set. MCVAE is trained in the variational framework by minimizing the negative evidence lower-bound (ELBO) in equation \ref{ELBO} with a Gaussian posterior $q_\phi$ (mean and co-variance parameterized from $(\Theta_M, \Theta_R)$) and the reconstruction loss $\mathcal{L}_M$ defined by Eq. (\ref{eq:match_loss}).
    
    {
        \small
        \begin{align}
             \ell_{ELBO}=KL(q_{\phi}||p(z))-\mathcal{L}_M(\Theta_R|\Theta_{R'}) \label{ELBO}
        \end{align}
    }
    We extend the Matching and MCVAE models to a multi-lingual setting by using pretrained multi-lingual BERT (MBERT) \cite{devlin2018bert} for $[f_{\varphi_M}, f_{\varphi_R}]$ similar to \cite{QiYingSRNAACL2021} and jointly training the models for all languages. 
        

    \section{\m: A Conditional Generative Matching Model for Reply Suggestion}\label{sec:model_description}
    Our initial analysis with universal models (jointly training models with all languages), reveals that the universal MCVAE performs better than Matching. However, simply training models jointly is not sufficient to achieve a models with high performance. First, the highly imbalanced nature of multi-lingual data leads to over- or under-fitting across languages resulting in performance worse than separately trained mono-lingual models. Second, training multi-lingual MCVAE proved is due to the reliance on a pretrained Matching model: it is not clear how to find a suitable Matching model checkpoint for initializing the MCVAE. Finally, since the text encoders for MCVAE are frozen during training, there is limited cross lingual transfer and improvement for low resource languages. Unfreezing the layers led to divergence of the model.
    
    To address the limitations of MCVAE, we propose an enhanced Conditional Generative Matching (\m) model, for the retrieval based RS  with inductive biases for the multi-lingual data and effective training techniques for creating high quality universal models.

    \subsection{Message Conditional Prior}
    The implied generative process in MCVAE (Fig. \ref{fig:mcvae_model}), is
    $p(z) \to p(\Theta_M|z) \to p(\Theta_{R}|\Theta_M,z)$, where the latent prior $z$ is sampled independent of the message encoding $\Theta_M$. However, in RS since $\Theta_M$ is always observed, ideally one would like to sample from $p(z|\Theta_M)$ to capture message-dependent information as well as rich multi-modality of the input space, particularly for multi-lingual data. In addition, although MCVAE works well empirically in the mono-lingual setting~\cite{Deb2019DiversifyingRS}, the
    samples from $p(z)$ in general are not the same as $p(z|\Theta_M)\propto p(z)p(\Theta_M|z)$, unless $p(\Theta_M|z)$ is uniform across the space of $\Theta_M$. This is a restrictive assumption, which motivates us to consider a prior conditioned on the input $\Theta_M$ for the generative model, by decomposing 
    
    {\small
        \begin{align}
            p(\Theta_{R},z|\Theta_M)=p(z|\Theta_M)p(\Theta_{R}|\Theta_M,z)
        \end{align}
    }
    as shown in Figure \ref{fig:cgm_model}. The conditional prior $p(z|\Theta_M)$
    is posed to encode message dependent information which can facilitate
    matching more relevant and diverse set of responses. 
    We define the message-conditional prior $p(z|\Theta_M)=\mathcal{N}(\mu(\Theta_M),\Sigma(\Theta_M))$, where the prior parameters are learnt from data during training and used for prediction, to maximally capture the multiple modalities of intents and intrinsically complex distribution of multi-lingual data.

    \subsection{Prior with Mixture Density (CGM-M)}
        We postulate that a more expressive conditional prior, such as a mixture density, can better capture the multi-lingual data in contrast to the single prior density as used above. I.e., the different components of a mixture density can represent different languages and allow independent representation across languages. To this end we extend the message conditional prior with a Gaussian Mixture model (GMM) as,
        
        {\small
            \begin{align}
              p(z|\Theta_M)=\sum_{k=1}^K \pi_k(\Theta_M) \mathcal{N}\left(\mu_k(\Theta_M), \Sigma_k(\Theta_M)\right)
            \label{eq:gaussian_mixtures}              
            \end{align}
        }
        where $\mu_k(\Theta_M)$, $\Sigma_k(\Theta_M)$ are the message dependent means and diagonal covariances for the $k$th component of the GMM, and $\pi_k(\Theta_M)$ are the message dependent prior mixing coefficients. We hypothesize that components would correspond to different intents and languages, thus providing additional inductive bias for multi-lingual data. We refer to the mixture variant as CGM-M (Figure \ref{fig:cgm_gmm_model}).

    \subsection{Aligning Latent Space to Language}
    To further reinforce the notion that the CGM-M latent components encode language specific information from M-R pairs, we pose an additional constraint that the language of the message be inferred from the prior mixture coefficient. This is instantiated by building a simple classifier network with loss $\ell_{LC}(l|\Theta_M, \pi)$ to map the prior mixture coefficient $\pi(\Theta_M)$ onto the language $l$ of the message.  We also tested with mapping the 1) means and variances $[\mu_k(\Theta_M), \Sigma_k(\Theta_M)]$, and 2) samples $z_k$ of the GMM, and found that mapping the $\pi(\Theta_M)$ leads to the best results. The classifier is learned jointly with the rest of the components.
                

    \subsection{Variational Training Architecture}
    \label{sec:model_learning}
    
    The CGM models are formulated as a VAE in the continuous space of $\Theta_M, \Theta_R$. \m includes two multi-lingual text encoders $[f_{\varphi_M}, f_{\varphi_R}]$, to convert the raw text of M-R into the common encoding space (encoders may be considered extraneous to the VAE but are learnt jointly with VAE layers), and a VAE with prior, posterior, and  generation networks $[p_{\psi}(\mu, \Sigma), q_{\phi}(\mu, \Sigma), g_{\theta}]$. 
    
    The \mm extends the \m version with category specific Gaussian components $[p_{\psi_c}, q_{\phi_c}]$
    In addition it also includes a categorical prior and posterior $[\pi_c, \rho_c]$, and a language classifier $l_c$ to discriminate between languages. We use the standard reparameterization trick for the Gaussian variables and the Gumbel-Softmax trick ~\cite{GumbelSoftmax} with hard sampling for the categorical variable. CGM-M (CGM is a special case with $K=1$) is summarized as follows.
    
    {\small
            \begin{align}
            &\textbf{Generative Model}:p_{\psi}(\mu, \Sigma), g_{\theta}\nonumber\\
                \pi &= \text{Softmax}(\texttt{FFN}_1(\Theta_M))\\
                c&=\text{GumbelSoftmax}(\texttt{FFN}_1(\Theta_M))\\
                \mu_\phi&=\texttt{FFN}_2(\Theta_{M}),\,\,
                \Sigma_\phi=\text{Softplus}(\texttt{FFN}_3(\Theta_{M}))\\
                 z_c&=\mu_{\phi_c}+\varepsilon\Sigma_{\phi_c},\; \text{where} \; \varepsilon\sim\mathcal{N}(0,I)\\
                 \Theta_{R'}&=
                 \texttt{FFN}_4(\overleftrightarrow{z_c\Theta_{M}})\\
                &\textbf{Variational Posterior}: q_{\phi}(\mu, \Sigma)\nonumber\\
                \rho &=\text{Softmax}(\texttt{FFN}_5(\overleftrightarrow{\Theta_{M}\Theta_{R}}))\\
                v&=\text{GumbelSoftmax}(\texttt{FFN}_5(\overleftrightarrow{\Theta_{M}\Theta_{R}}))\\
                \mu_{\psi}&=\texttt{FFN}_6(\overleftrightarrow{\Theta_{M}\Theta_{R}})\\
                \Sigma_{\psi}&=\text{Softplus}(\texttt{FFN}_7(\overleftrightarrow{\Theta_{M}\Theta_{R}}))\\
                 z_v&=\mu_{\psi_v}+\xi\Sigma_{\psi_v},\; \text{where} \; \xi\sim\mathcal{N}(0,I)
            \end{align}
        }
    Above, we expand the dimensions of projection vectors to $\mu:[h \times K],\Sigma:[h \times K]$ where $h$ is the dimension of the forward projections and $K$ is the number of categories in the mixture. After the category is selected (using Gumbel Softmax), we use the category index to select part of the expanded projections, as the $k^{th}$ component of the means and variances $(\mu_k, \Sigma_k)$. Each $\texttt{FFN}_i$ denotes a two-layer feed-forward network (except $\texttt{FFN}_4$ which has 3 layers) with \texttt{tanh} activation and $\leftrightarrow$ denotes vector concatenation. 
    
    Note that the posteriors are conditioned on both $\Theta_M$ and $\Theta_R$. This theoretically provides a richer representation of the M-R pairs and during inference allows us to score the combination of message and the selected response vectors. However, during training, it can lead to leakage through the network where the model simply ignores the message and uses the response vector for generation. We mitigate the leakage by applying a low-dimensional projection of response vector $\Theta_R$ before feeding into the variational network.

    Following standard stochastic gradient variational bayes (SGVB) training, we minimize the negative ELBO to train the network. \mm adds the classifier loss to enforce alignment between latent vectors and language types. The training objectives for each are given as follows,
    
    {\small
    \begin{align}
      \ell_{\text{\m}} &= KL(q_{\phi}||p_{\psi}) - \mathcal{L}(\Theta_R|\Theta_{R'})\\
        \ell_{\text{\mm}} &= KL_{M}(q_{\phi}||p_{\psi}) - \mathcal{L}(\Theta_R|\Theta_{R'}) +  \ell_{LC}
    \end{align}}
    where the reconstruction log-loss, $\mathcal{L}(\Theta_R|\Theta_{R'})$ is given by Eq. (\ref{eq:match_loss}). For CGM, the KL divergence between the two multivariate Gaussian densities can be computed in closed form. However, for CGM-M, the KL divergence between two Gaussian mixtures does not admit a closed form. We estimate it with a variational approximation method described in \cite{KLD-approx-GMM}\footnote{Another approach with Monte-Carlo sampling requires a large number of samples and was not as effective.}.
    
    {\small
        \begin{align}
            KL_{M}(q||p) &\approx \sum_{i=1}^K \pi_i \log \frac 
            {\sum_{j=1}^K \pi_j e^{KL(p_{\phi_i}||q_{\psi_j})} } 
            {\sum_{k=1}^K \rho_k e^{KL(p_{\phi_i}||q_{\psi_k})} }
        \end{align}
    }
    \subsection{Training Optimizations} \label{sec:training_optimizations}
    Training deep generative models with SGVB has been known to be notoriously tricky \cite{SamBow16, fu-etal-2019-cyclical}. Our multilingual setting, and joint training of text encoders with VAE layers makes it even more challenging. We employed several optimizations to improve the convergence of the models.
    
    \textbf{1) Matching loss regularization}:
    In CGM, the encoders for $\Theta_M, \Theta_R$ are learnt jointly with the VAE layers in order to maximize richness of shared latent representation across languages. Thus $\Theta_R$ is a moving target for the VAE generator outputting $\Theta_{R'}$ and causes the training to diverge without additional constraints.
    In MCVAE, this was mitigated by initializing and freezing the text encoders from a trained Matching model, but can be counter-productive in the multilingual scenario. To enable joint training of text encoders and the VAE layers, and mitigate the issue of a moving target for reconstruction, we introduce a regularization in the form of a matching score between $\Theta_M$ and $\Theta_R$,
    
    {\small
      \begin{equation}
        \ell_{\text{\mm}} = KL_{M}(q_{\phi}||p_{\psi}) - \mathcal{L}(\Theta_R|\Theta_{R'}) +  \ell_{LC}
        -\mathcal{L}(\Theta_R|\Theta_M)
      \end{equation}
    }which constrains the response vector to have a representation close to the message vector. This provides an independent anchor for the reconstruction and allows the end-to-end training of the model utilizing the full parameter space of the encoders for enhanced representation.
    
    \textbf{2) Multi-sample variance scaling}:
    In SGVB,  using a single sample of $z$ usually results in high variance in the ELBO estimate. One remedy is to estimate the ELBO with multiple samples, either in the non-weighted and or importance weighted~\cite{IWAE} versions. However, these led to only minor improvements.
        

                


    In multi-sample training we take the expectation of the ELBO over the samples. We found that if instead we first take the expectation of the samples $z'=\sum_{i=1}^k z_i/k$ before computing the ELBO loss, we can reduce the variance and stabilize the training. Since $z'$ follows an equivalent distribution $z'\sim \mathcal{N}(\mu,\frac{\Sigma}{k})$, we can estimate ELBO with multiple samples drawn from the scaled distribution and compute the expectation as follows. The adjustment provides significant improvements in training convergence and metrics.

    {\small
      \begin{gather}
        \ell_{CGM}=\mathbb{E}_{z'}[-KL_{z'}(q_{\phi}||p_{\psi}) + \mathcal{L}(\Theta_R|\Theta_{R'})] 
        \label{eq:elbo}
      \end{gather}
    }
    \textbf{3) Weighting loss components with Homo-scedastic Uncertainty (HSU)}: 
    The final loss formulations for both \m and \mm have several components. For finer control of training, we introduce learnable weights $w_i$ for each of the components. Weighting different components of the ELBO loss has shown to improve performance \cite{beta-VAE} in SGVB and thus even without additional components, such a weighting process is recommended.
    
    Following \cite{HSU_Cipolla_Gal}, we view the loss formulation as a multi-task learning objective with different \textit{homo-scedastic} uncertainties (HSU) for each task. Assuming the components factorize to Gaussian (continuous) and discrete (cross-entropy) likelihoods, the loss with HSU can be viewed as:
        
    {\small
        \begin{equation}
            \begin{split}
                \ell_{HSU}=&\frac{1}{2\sigma_1^2} KL(q_{\phi}||p_{\psi}) - \frac{1}{2\sigma_2^2} \mathcal{L}(\Theta_R|\Theta_{R'})\\
                &- \frac{1}{2\sigma_3^2} \mathcal{L}(\Theta_R|\Theta_M) +\frac{1}{2\sigma_4^2}\ell_{LC}\\
                &+ \log(\sigma_1) + \log(\sigma_2) + \log(\sigma_3) + \log(\sigma_4)
            \end{split}
        \end{equation}
    }
    Equating the uncertainties with the weights in our loss equation, this can be seen as learning the relative weights for each component where $w_i \sim 1/\sigma_i^2$ and provides a smooth, regularized and differentiable interpretation of weights. We introduce the weights as parameters in the model and learn them jointly with rest of the network.
        
    \textbf{4) Handling data skew with Focal Loss (FL)}:
    Multilingual training  can have different convergence rates across languages and akin to behaviors observed in multi-modal training \cite{gradient-vaccine}.  Carefully configured sampling ratios for different languages can alleviate this problem but requires costly hyper-parameter search. Instead we employ a popular technique for handling skewed data distribution: the focal loss (FL) \cite{focal_loss}. 
    
    {\small
        \begin{equation}
            \begin{split}
                \mathcal{L}_{FL}(\Theta_R|\Theta_{R'})= (1- e^{\mathcal{L}(\Theta_R|\Theta_{R')}})^\alpha \mathcal{L}(\Theta_R|\Theta_{R'})
            \end{split}
        \end{equation}
    }
    The FL (with $\alpha=1$) is applied on the reconstruction log-probability component of ELBO, such that strongly reconstructed vectors are given lower weights than the weakly reconstructed ones which balances the convergence across languages. 
    
    \subsection{Prediction and Ranking Responses}
    
    During prediction, we rank and select responses from a fixed response set $R_{[s]}$. Since the models generate response vectors in the continuous space, the prediction process needs to convert the samples into ranking in the discrete space of responses. The process is described as follows.
    
    {\small
            \begin{align}
                &\log{p_i(\Theta_{R_{[s]}}|\Theta_M)}= \mathcal{L}(\Theta_{R_i'}|\Theta_{R_{[s]}}) - KL_z(q||p)\label{eq:pred_scores}\\ 
                 &MRR(R_{[s]})= \frac{1}{N} \sum_i^N [Rank_{R_{[s]}}\log{p_i(\Theta_{R_{[s]}}|\Theta_M)}]^{-1} \label{eq:MRR}
            \end{align}
    }
    For each message we generate 1000 samples of latent conditional priors from $z\sim\mathcal{N}(\mu_{\phi},\Sigma_{\phi})$ and from categorical prior for CGM-M. Next, we generate samples of the response vectors using the generator network, $\Theta_{R_i'} \sim g_{\theta}(\Theta_{R_i'}|\Theta_M,z_i)$. We compute the scores for the $i^{th}$ generated sample w.r.t to the fixed response set $\log{p_i(\Theta_{R_{[s]}}|\Theta_M)}$ in eq. \ref{eq:pred_scores}, where the KL divergence is directly computed on the samples $z$ under a Normal or GMM distribution for the prior and posterior. To reduce the scoring overhead over 40k responses with 1000 samples, we pre-select top $k$ ($k=100$ provides sufficiently diverse candidates) using the matching score (eq. \ref{eq:match_score}). Finally, the mean reciprocal ranks (MRR) over all the samples (eq. \ref{eq:MRR}) are used to select the top 3 as our predicted responses.
         \begin{table*}[htbp!]
          \centering
          {\small
          \begin{tabular}{cccccc}\toprule
            & Latent Factors & Cond. Prior & Mix. Density & Language alignment & Multilingual training opts \\
            \midrule
                Matching & -& -& -& -& -\\
                MCVAE & \checkmark& -& -& -& -\\
                \m & \checkmark& \checkmark& -& -& \checkmark \\
                CGM-M & \checkmark& \checkmark& \checkmark&\checkmark&\checkmark \\
            \bottomrule
          \end{tabular}
          }
          \caption{Comparison of components of Matching, MCVAE (Sec \ref{sec:background}), CGM, and CGM-M (Sec \ref{sec:model_description})} \label{modelsummary}
      \end{table*}
      
    \begin{figure*}[t]
    	\centering
    	\includegraphics[trim={0 80 0 80},clip,width=0.49\textwidth]{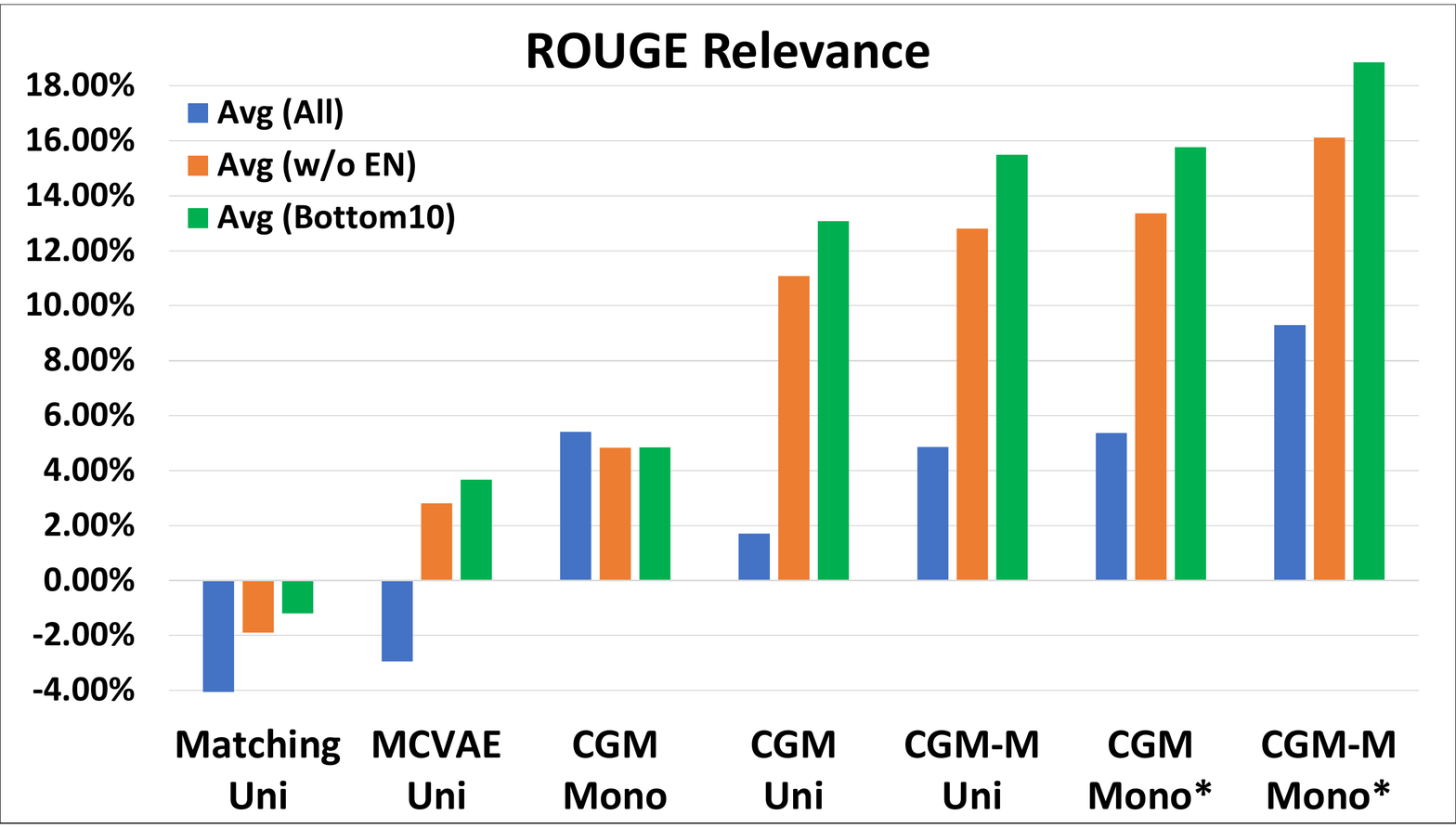}
        \includegraphics[trim={0 80 0 80},clip,width=0.49\textwidth]{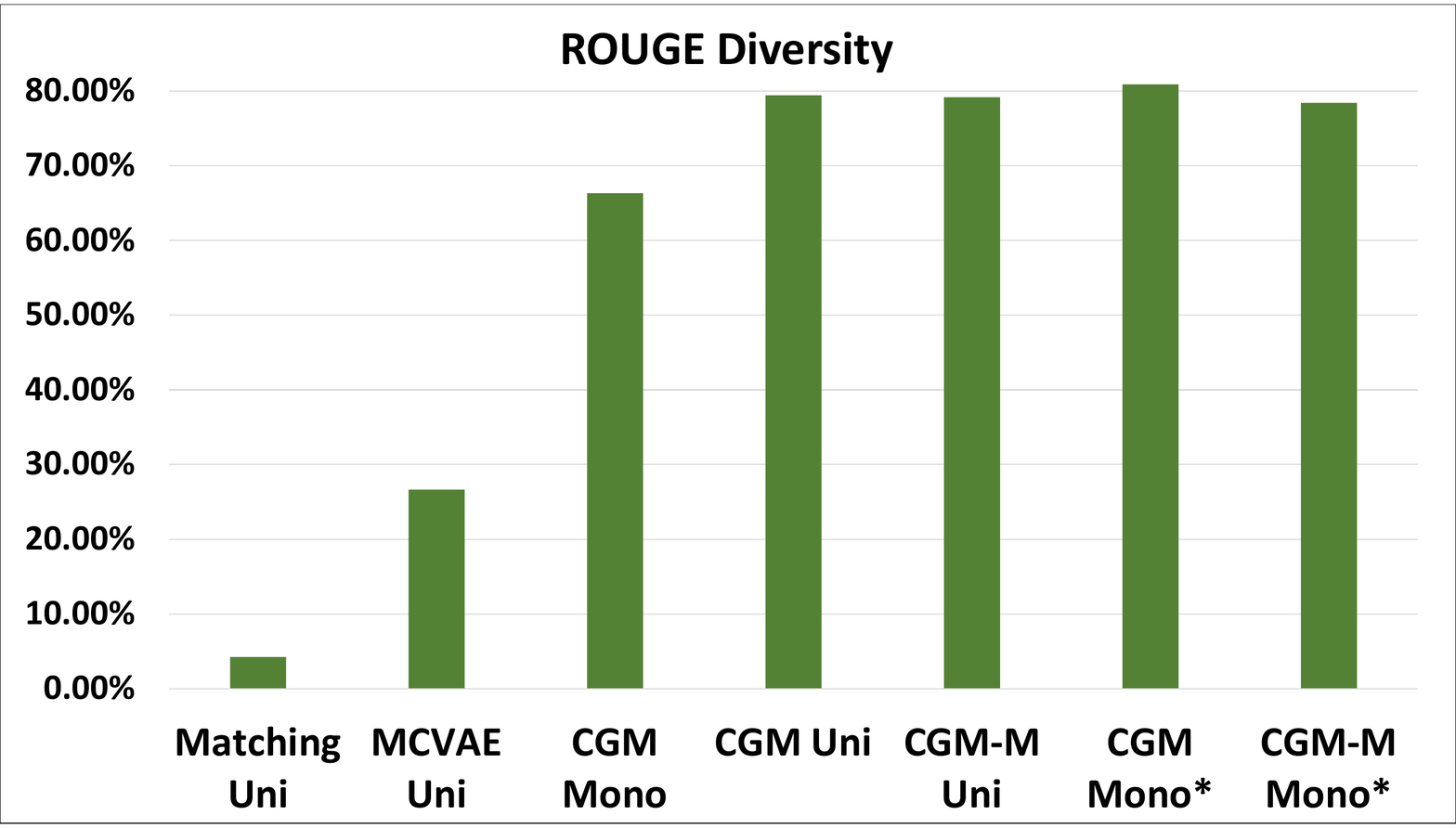}
    	\caption{Main results. With the Matching monolingual models as baseline, the figures show the \% changes in metrics for  model variants (see Sec \ref{sec:model_variants} for model description and Sec \ref{sec:main_results} for discussion). For each model variant, we show the metrics across three languages groups (All, w/o-EN and bottom 10 low resource languages.  (Left) Relevance (Right) Diversity.}
    	\label{tab:OverallBarCharts}
    \end{figure*}

\section{Experiments}

\begin{figure*}
    	\centering
    	\includegraphics[scale=0.56,trim={0 250 0 80},clip]{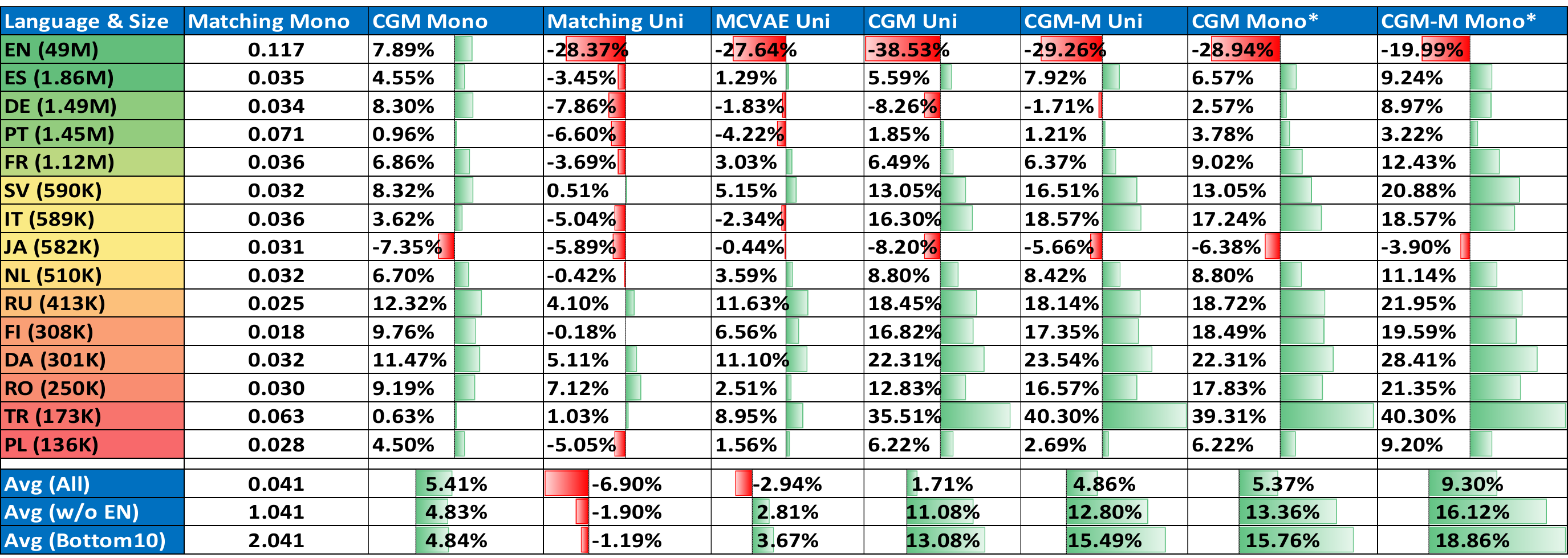}
    	\caption{Relevance metrics across 15 languages. (Model description in Sec \ref{sec:model_variants} and discussion in Sec \ref{sec:all_languages_results})
    	}
    	\label{tab:MonolingualTable}
    \end{figure*}

\textbf{Multi-lingual data}:
We use the MRS (Mulit-lingual Reply Suggestions) data set \cite{zhang-etal-2021-dataset} for our experiments. MRS consists of message-reply (M-R) pairs separated into different languages from Reddit conversations \cite{baumgartner2020pushshift} using the FastText detector~\citep{joulin2016fasttext}. We select the top 15 languages for experimentation (data volume was insufficient for others) with 80\% split for training (2nd column in Table \ref{tab:MonolingualTable}) and the rest for validation and test. We create response sets with most frequent responses (>20 frequency) in the m-r pairs. For low resource languages, we augment this natural set with machine translated responses from EN, resulting in $\sim40$k responses for each language.

\textbf{Metrics}: 
We use ROUGE~\cite{lin-2004-rouge} for scoring the \textit{relevance} of the 3 predicted responses against the reference response.  We also compute the self-ROUGE~\cite{celikyilmaz2020evaluation} within the 3 responses as a measure of \textit{diversity}. For both, we report the average of the ROUGE-F1 for 1/2/3-grams across the three responses.

\textbf{Train parameters}: 
We use the multi-lingual version of the pretrained BERT model (MBERT) \cite{devlin2018bert} as out text encoders for which we use the Huggingface's transformers library \cite{wolf-etal-2020-transformers}. We freeze the embedding layer of MBERT encoders, which reduces training overhead, and preserves cross-lingual representation without impacting performance \cite{lee2019would, peters2019tune}. We use dimension size of 512 for the VAE layers. For CGM-M we set the number of categories to $K=20$.

We train with the Adam optimizer (peak rate: $1e-5$, exp. decay: $0.999$ after warm up of 1000 steps), batch size of 256, and m-r pairs truncated to length 64 and 32 respectively. We add language tokens (e.g. EN, PT) before m-r pairs as additional language identifier. All the model sizes are relatively similar (1.3GB to 1.5GB) since most parameters are in the two MBERT encoders with 12 transformer layers (each around 700MB).

\textbf{Multilingual training}: 
We uniformly sample languages such that models have equal exposure to each language during training. This leads to good performance across all languages except EN. Alternatively, sampling proportionate to data volumes, had good performance for EN but led to severe under-fitting for most languages other than EN as EN dominates the training with orders of magnitude more data. The ideal sampling is somewhere in between, but requires extensive search to optimize. On single NVidia V100 GPUs, models converge within 1-2 epochs $\sim48hrs$ over the entire data (i.e., 1-2 epochs for EN and multiple epochs for others). Joint training amortizes the training costs, and can be used even when targeting monolingual models, by saving per-language checkpoints. 

\phantomsection
\label{sec:model_variants}
\textbf{Model variants}: We analyze 4 models: Matching, MCVAE, CGM and CGM-M (Table \ref{modelsummary}). For each we consider 3 multilingual model variants.  \textbf{[Mono]}: individually trained monolingual models on each language. \textbf{[Uni]}: jointly trained universal model with a single checkpoint for evaluation. \textbf{[Mono*]}: jointly trained model with per language checkpoints (saved when the validation metrics peak for each language) for evaluation. Since models peak at different point for each language, Mono* is expected to have a better performance than the Universal counterpart with a single checkpoint. 



\subsection{Main Results}\label{sec:main_results}
Figure \ref{tab:OverallBarCharts} shows the relevance and diversity metrics for different model variants. With Matching-Mono models (trained individually per language) as the baseline, we plot the \% changes in metrics for the other model variants. Models are trained on all languages, with relevance metrics shown in 3 language groups: 1) All, 2) All w/o EN, and 3) Bottom 10 low resource languages, to highlight the differences from data volumes in languages.\footnote{Here we present quantitative results. For qualitative analysis, multi-lingual text predictions are provided in the appendix.}

\textbf{Relevance} (Figure~\ref{tab:OverallBarCharts}-Left): Compared to individually trained monolingual Matching model, the universally trained Matching-Uni regresses on all the three language group while MCVAE-Uni improves for latter two groups (w/o EN and bottom 10 languages). The CGM-Mono improves the metrics across all three languages. Thus even without joint training, CGM by itself is better than the baselines and thus raises the bar which the universal models needs to match or overcome. 

The CGM and CGM-M universal models improve on all the language groups although for the CGM-uni, there is regression in the \textit{All}-languages group compared to the CGM-mono (more discussion later). However, CGM-M-Uni with around 5\% increase is actually slightly better than CGM-mono, showing that we can replace the monolingual models with a single universal model. Next, the Mono* models (universally trained but with best per-language checkpoints saved) can achieve even bigger gains and CGM-M-Mono* surpasses other models in every language group. 

Within language groups, we observe increase upto 16\% without EN and upto 19\%  for bottom 10 languages. EN with two orders of magnitude more data, remains severely under-fitted in all the jointly trained model, due to which the metrics improvements in \textit{All} languages group remains low. 

\textbf{Diversity} (Figure \ref{tab:OverallBarCharts}-Right): The CGM performance is most striking for diversity metrics where we see 80\% improvements. Diversity improvements more than the relevance gains, illustrate that deep generative modeling enhancements in CGM leads to richer representation of multilingual data
with improved discrimination and disentanglement of language and latent intents in M-R pairs. 
CGM-M achieves high diversity on top of the best relevance metrics, showing the enhanced representation through mixture models. 

    \begin{figure}
    	\centering
    	\includegraphics[scale=0.270,trim={0 130 0 80},clip]{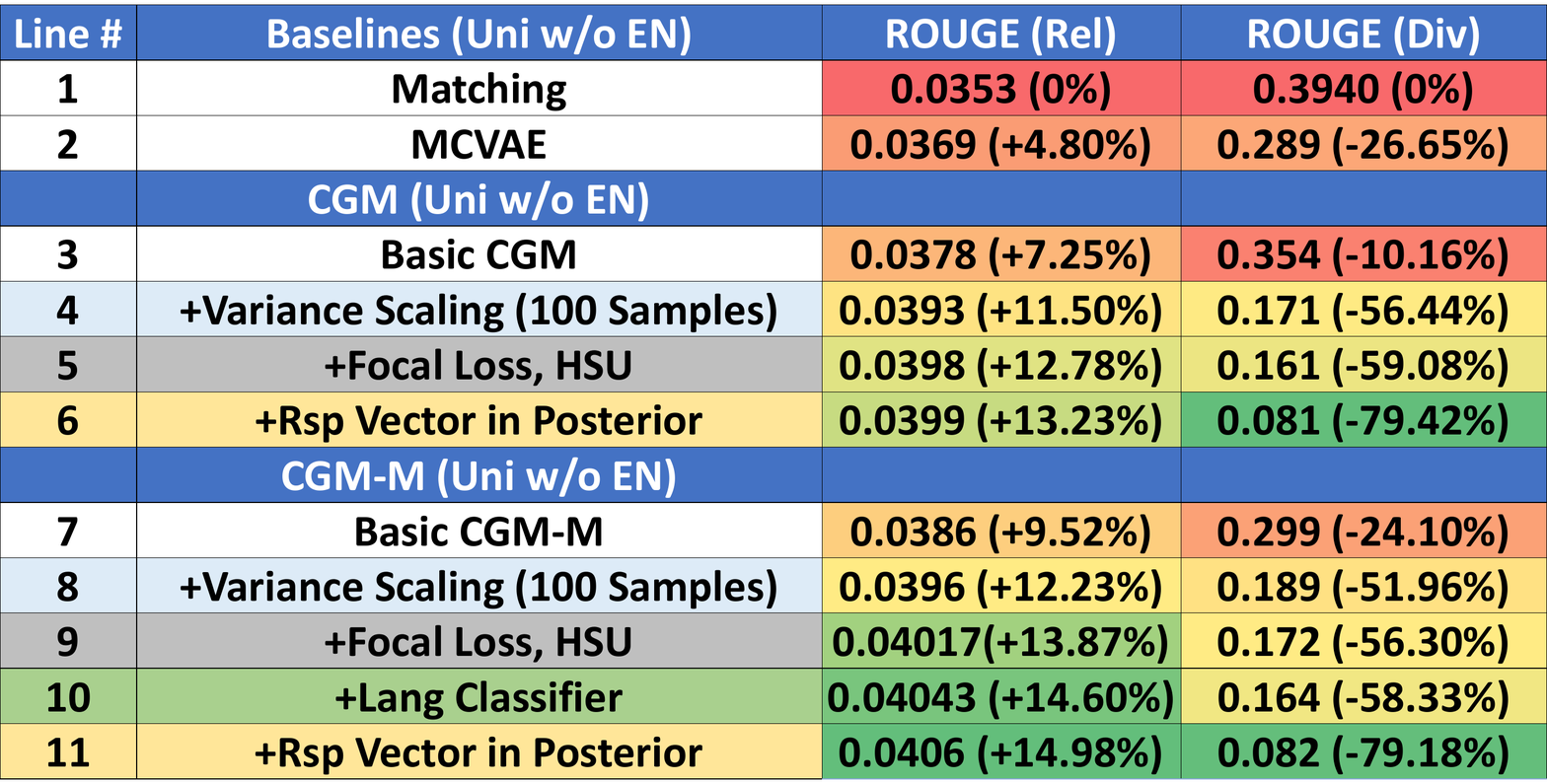}
    	\caption{Ablation studies for different training optimizations (Sec \ref{sec:training_optimizations}) with results discussed in Sec \ref{sec:ablation_studies}.}
    	\label{tab:AblationsTable}
    \end{figure}
    
\subsection{Ablation Studies}\label{sec:ablation_studies}
We conducted extensive ablation studies with the different model variants, and training optimizations and summarize the results in Figure \ref{tab:AblationsTable}. For ablations we report the metrics for language group without EN, as the significantly higher data volume in EN can conflate the results. 

\textbf{Baselines}: We use the Matching-uni model (line 1) as the baseline. MCVAE (line 2) improves both relevance (4.8\%) and diversity (27\%) which shows the potential of deep generative models.

\textbf{Training optimizations with CGM}: The basic CGM-Uni model (line 3) and CGM-M (Line 7) shows modest relevance gains compared to MCVAE. We attribute the modest gains due to complexities with end-to-end training of the CGM. Through training optimizations of variance scaling, and FL and HSU (lines 4, 5), CGM can comfortably surpass MCVAE in relevance (12.8\%) and double the diversity (59\%). CGM-M, shows similar increase (13.87\%) with variance scaling (line 8), and FL and HSU (line 9) outperforming the best achieved with CGM. The biggest improvements come from  multi-sample variance scaling (lines 4, 8) with additional improvements from FL and HSU (lines 5, 9). Overall, the optimizations lead to more stable training, and faster convergence across languages. They also alleviate the need for manual tuning for skewed data and loss component weights, making the training process virtually hyper-parameter free.

\textbf{Language Mapping in CGM-M:} 
One key reason for improved performance with CGM-M is the potential inductive bias for languages through the mixture components, which can be further boosted by explicit mapping of latent vectors to languages. 
Language mapping improves the relevance to 14.6\% (line 10) over the baseline. We also see a slight boost in diversity showing the improved modeling of the multi-lingual distribution using this approach.


\textbf{Posterior conditioned on both message and response}: 
The joint conditioning of the posterior with both the $\Theta_M, \Theta_R$  vectors\footnote{We had excluded $\Theta_R$ in the posterior of other configurations to show this effect.} gives the best relevance for both CGM and CGM-M (lines 6, 11) with CGM-M exceeding all other variants. More interesting is the substantial improvement in diversity (80\%), which illustrates that it encourages a richer representation in the prior by perhaps disentangling latent intents and language characteristics better. We note here that, in CGM-M, using the full $\Theta_R$ dimension (768) led to high level of leakage through the posterior (multiple components of the mixture further aids the leakage). We use a low dimensional projection of size 16 in CGM-M to mitigate the issue. 


\subsection{Analysis across Languages Groups}\label{sec:all_languages_results}

Next, we discuss the performance breakdown of models across individual languages. Figure \ref{tab:MonolingualTable} expands the Relevance metrics from Figure \ref{tab:OverallBarCharts} for all languages. As before, we use the the Matching-Mono as the baseline, and list the \% changes over this baseline for each model and language.



We see that, all jointly trained variants (Uni and Mono*) have severe under fitting for EN. In fact if we simply remove EN from the metrics the CGM variants vastly improve upon the monolingual versions. With almost two orders of magnitude more data in EN (49M), it remains challenging to have good performance simultaneously for EN and other languages without additional tricks. In general the improvements are less for the top 5 high-resource languages which can be attributed to lesser impact from information sharing and lower exposure of these languages due to uniform sampling. Such issues have been reported in prior literature as capacity dilution ~\citep{johnson2017google,conneau2020unsupervised,wang2020negative} where there is always a trade-off between low and high resource languages. CGM while not completely eliminating it, largely mitigates the issue.

The impact of CGM with joint training is more pronounced for the bottom 10 language group. For example we see 15.49\% improvement for CGM-M compared to only 3.67\% for MCVAE-Uni. Finally, we see improvements of 15.76\%  for CGM-Mono* and 18.86\% for CGM-M Mono* models, illustrating that even if we target mono-lingual models, CGM can take advantage of shared learning through joint training while saving compute.

The improvements for low resource languages, show that CGM is more data efficient due to model enhancements, while the prevention of regressions for high resource languages show a more balanced learning through training optimizations. The fact that these relevance improvements come in addition to 80\% improvements in diversity, shows the remarkable effectiveness of CGM to represent the multi-modal landscape of multi-lingual RS.

     \section{Related Work}
   
    VAEs have been used in retrieval based Q\&A \cite{crossing-VAE-QA}, document matching \cite{ChaidaroonSIGIR17}, and recommendations \cite{VAETopN}. CGM for RS is most closely related to MCVAE \cite{Deb2019DiversifyingRS} but differs in the expressive conditional priors, multi-component mixture density priors, language alignment, and training optimizations which makes it effective in a multi-lingual setting. 
    
    For multi-task scenarios, VAEs can offer significant modeling efficiencies \cite{cao2020modelling, CURL-GMVAE} with additional improvements through mixture model priors, e.g.  in \cite{VAE-GMM_clustering, VAE-GMM-Graphs} for unsupervised clustering, in \cite{Meta-GMVAE} for unsupervised meta-learning, and in \cite{Var-MOE} as a multi-modal variational mixture-of-experts.  
    
    VAEs can also improve multilingual representation for low resource languages, e.g. in models like BERT \cite{li2020optimus}, in \cite{VAE-CrossLingual} for document classification, in \cite{VAE-Speech} for disentangling phonemes for speech synthesis, and in \cite{VAE-MT, eikema-VAE-MT} for neural machine translation. VAEs can improve diversity in language generation and retrieval tasks \cite{Zhao2017LearningDD, Tran2017AGA, Shen2017ACV, Deb2019DiversifyingRS} through better modeling efficiencies. Such results motivated us to apply VAEs for multilingual RS. 
    
    We may also consider alternative to VAEs such as training auxiliary tasks with adapters \cite{adapters_houlsby19a}, adversarial learning~\citep{chen2018adversarial,chen2019multi,huang2019cross}, and mixing pre-training and fine-tuning~\citep{phang2020english} to improve modeling in multilingual setting. This is subject of future work. We also plan to experiment with higher capacity multilingual encoders such XLM-R \cite{lample2019cross} and InfoXLM \cite{chi2020infoxlm} to further improve the performance. However, the choice of the base encoder is orthogonal to the improvements (especially on diversification) shown in this paper.
    
    
    


    As noted in prior work, multilingual training can have capacity dilution issues
    ~\citep{johnson2017google,conneau2020unsupervised,wang2020negative}. Overall, multilingual models are closing the gap with monolingual counterparts for wide range of tasks \cite{QiYingSRNAACL2021, ranasinghe2020multilingual, MLUniEnc}, and as shown in this paper, even surpass them. Careful sampling strategies, and techniques such as Translation Language Model (TLM) can alleviate the "curse of multilinguality" ~\citep{lample2019cross} but we show improvements without additional data augmentation (translation pairs), and with simple uniform sampling.
    
    

    
    

    \section{Conclusions}
In this paper we present a conditional generative Matching model (CGM) for retrieval based suggested replies. CGM not only provides relevance gains (15\%), but also substantial improvements in diversity (80\%). While CGM clearly advances the state of art for modeling multi-lingual RS systems, it also 
illustrates that through proper model choices and training optimizations, we can surpass and replace monolingual models. This is important for both industry and academia and suggests similar strategies to be applied across diverse tasks.  This is subject of future work. 
    \bibliographystyle{acl_natbib}
    \bibliography{main_camera_ready}

\begin{thebibliography}{57}
\expandafter\ifx\csname natexlab\endcsname\relax\def\natexlab#1{#1}\fi

\bibitem[{Baumgartner et~al.(2020)Baumgartner, Zannettou, Keegan, Squire, and
  Blackburn}]{baumgartner2020pushshift}
Jason Baumgartner, Savvas Zannettou, Brian Keegan, Megan Squire, and Jeremy
  Blackburn. 2020.
\newblock The pushshift reddit dataset.
\newblock In \emph{Proceedings of the international AAAI conference on web and
  social media}, volume~14, pages 830--839.

\bibitem[{Bowman et~al.(2016)Bowman, Vilnis, Vinyals, Dai, Józefowicz, and
  Bengio}]{SamBow16}
Samuel~R. Bowman, Luke Vilnis, Oriol Vinyals, Andrew~M. Dai, Rafal Józefowicz,
  and Samy Bengio. 2016.
\newblock {Generating sentences from a continuous space}.
\newblock In \emph{CoNLL}.

\bibitem[{Burda et~al.(2016)Burda, Grosse, and Salakhutdinov}]{IWAE}
Yuri Burda, Roger Grosse, and Ruslan Salakhutdinov. 2016.
\newblock Importance weighted autoencoders.
\newblock In \emph{ICLR}.

\bibitem[{Cao and Yogatama(2020)}]{cao2020modelling}
Kris Cao and Dani Yogatama. 2020.
\newblock Modelling latent skills for multitask language generation.
\newblock \emph{arXiv preprint arXiv:2002.09543}.

\bibitem[{Celikyilmaz et~al.(2020)Celikyilmaz, Clark, and
  Gao}]{celikyilmaz2020evaluation}
Asli Celikyilmaz, Elizabeth Clark, and Jianfeng Gao. 2020.
\newblock Evaluation of text generation: A survey.
\newblock \emph{arXiv preprint arXiv:2006.14799}.

\bibitem[{Chaidaroon and Fang(2017)}]{ChaidaroonSIGIR17}
Suthee Chaidaroon and Yi~Fang. 2017.
\newblock Variational deep semantic hashing for text documents.
\newblock In \emph{Proceedings of the 40th International ACM SIGIR Conference
  on Research and Development in Information Retrieval}, pages 75--84.

\bibitem[{Chen et~al.(2019)Chen, Hassan, Hassan, Wang, and
  Cardie}]{chen2019multi}
Xilun Chen, Ahmed Hassan, Hany Hassan, Wei Wang, and Claire Cardie. 2019.
\newblock Multi-source cross-lingual model transfer: Learning what to share.
\newblock In \emph{Proceedings of the 57th Annual Meeting of the Association
  for Computational Linguistics}, pages 3098--3112.

\bibitem[{Chen et~al.(2018)Chen, Sun, Athiwaratkun, Cardie, and
  Weinberger}]{chen2018adversarial}
Xilun Chen, Yu~Sun, Ben Athiwaratkun, Claire Cardie, and Kilian Weinberger.
  2018.
\newblock Adversarial deep averaging networks for cross-lingual sentiment
  classification.
\newblock \emph{Transactions of the Association for Computational Linguistics},
  6:557--570.

\bibitem[{Chen and de~Rijke(2018)}]{VAETopN}
Yifan Chen and Maarten de~Rijke. 2018.
\newblock A collective variational autoencoder for top-n recommendation with
  side information.
\newblock In \emph{Proceedings of the 3rd Workshop on Deep Learning for
  Recommender Systems}, pages 3--9.

\bibitem[{Chi et~al.(2021)Chi, Dong, Wei, Yang, Singhal, Wang, Song, Mao,
  Huang, and Zhou}]{chi2020infoxlm}
Zewen Chi, Li~Dong, Furu Wei, Nan Yang, Saksham Singhal, Wenhui Wang, Xia Song,
  Xian-Ling Mao, He-Yan Huang, and Ming Zhou. 2021.
\newblock Infoxlm: An information-theoretic framework for cross-lingual
  language model pre-training.
\newblock In \emph{Proceedings of the 2021 Conference of the North American
  Chapter of the Association for Computational Linguistics: Human Language
  Technologies}, pages 3576--3588.

\bibitem[{Chorowski et~al.(2019)Chorowski, Weiss, Bengio, and van~den
  Oord}]{VAE-Speech}
Jan Chorowski, Ron~J. Weiss, Samy Bengio, and Aaron van~den Oord. 2019.
\newblock \href {https://dl.acm.org/doi/10.1109/TASLP.2019.2938863}
  {Unsupervised speech representation learning using wavenet autoencoders}.
\newblock In \emph{IEEE/ACM Transactions on Audio, Speech, and Language
  Processing}.

\bibitem[{{Cipolla} et~al.(2018){Cipolla}, {Gal}, and
  {Kendall}}]{HSU_Cipolla_Gal}
R.~{Cipolla}, Y.~{Gal}, and A.~{Kendall}. 2018.
\newblock \href {https://doi.org/10.1109/CVPR.2018.00781} {Multi-task learning
  using uncertainty to weigh losses for scene geometry and semantics}.
\newblock In \emph{2018 IEEE/CVF Conference on Computer Vision and Pattern
  Recognition}, pages 7482--7491.

\bibitem[{Conneau et~al.(2020)Conneau, Khandelwal, Goyal, Chaudhary, Wenzek,
  Guzm{\'a}n, Grave, Ott, Zettlemoyer, and Stoyanov}]{conneau2020unsupervised}
Alexis Conneau, Kartikay Khandelwal, Naman Goyal, Vishrav Chaudhary, Guillaume
  Wenzek, Francisco Guzm{\'a}n, Edouard Grave, Myle Ott, Luke Zettlemoyer, and
  Veselin Stoyanov. 2020.
\newblock \href {https://doi.org/10.18653/v1/2020.acl-main.747} {Unsupervised
  cross-lingual representation learning at scale}.

\bibitem[{Deb et~al.(2019)Deb, Bailey, and Shokouhi}]{Deb2019DiversifyingRS}
Budhaditya Deb, P.~Bailey, and M.~Shokouhi. 2019.
\newblock Diversifying reply suggestions using a matching-conditional
  variational autoencoder.
\newblock In \emph{NAACL-HLT}.

\bibitem[{Devlin et~al.(2019)Devlin, Chang, Lee, and
  Toutanova}]{devlin2018bert}
Jacob Devlin, Ming-Wei Chang, Kenton Lee, and Kristina Toutanova. 2019.
\newblock Bert: Pre-training of deep bidirectional transformers for language
  understanding.
\newblock In \emph{Proceedings of the 2019 Conference of the North American
  Chapter of the Association for Computational Linguistics: Human Language
  Technologies, Volume 1 (Long and Short Papers)}, pages 4171--4186.

\bibitem[{Dilokthanakul et~al.(2017)Dilokthanakul, Mediano, Garnelo, Lee,
  Salimbeni, Arulkumaran, and Shanahan}]{VAE-GMM_clustering}
Nat Dilokthanakul, Pedro A.~M. Mediano, Marta Garnelo, Matthew C.~H. Lee, Hugh
  Salimbeni, Kai Arulkumaran, and Murray Shanahan. 2017.
\newblock \href {http://arxiv.org/abs/1611.02648} {Deep unsupervised clustering
  with gaussian mixture variational autoencoders}.

\bibitem[{Eikema and Aziz(2019)}]{eikema-VAE-MT}
Bryan Eikema and Wilker Aziz. 2019.
\newblock \href {https://doi.org/10.18653/v1/W19-4315} {Auto-encoding
  variational neural machine translation}.
\newblock In \emph{Proceedings of the 4th Workshop on Representation Learning
  for NLP (RepL4NLP-2019)}, pages 124--141, Florence, Italy. Association for
  Computational Linguistics.

\bibitem[{Fu et~al.(2019)Fu, Li, Liu, Gao, Celikyilmaz, and
  Carin}]{fu-etal-2019-cyclical}
Hao Fu, Chunyuan Li, Xiaodong Liu, Jianfeng Gao, Asli Celikyilmaz, and Lawrence
  Carin. 2019.
\newblock \href {https://doi.org/10.18653/v1/N19-1021} {Cyclical annealing
  schedule: A simple approach to mitigating {KL} vanishing}.
\newblock In \emph{Proceedings of the 2019 Conference of the North {A}merican
  Chapter of the Association for Computational Linguistics: Human Language
  Technologies, Volume 1 (Long and Short Papers)}, pages 240--250, Minneapolis,
  Minnesota. Association for Computational Linguistics.

\bibitem[{Henderson et~al.(2017)Henderson, Al-Rfou, Strope, Sung, Luk{\'a}cs,
  Guo, Kumar, Miklos, and Kurzweil}]{henderson2017efficient}
Matthew Henderson, Rami Al-Rfou, Brian Strope, Yun-Hsuan Sung, L{\'a}szl{\'o}
  Luk{\'a}cs, Ruiqi Guo, Sanjiv Kumar, Balint Miklos, and Ray Kurzweil. 2017.
\newblock Efficient natural language response suggestion for smart reply.
\newblock \emph{arXiv preprint arXiv:1705.00652}.

\bibitem[{Henderson et~al.(2019)Henderson, Vuli{\'c}, Gerz, Casanueva,
  Budzianowski, Coope, Spithourakis, Wen, Mrk{\v{s}}i{\'c}, and
  Su}]{Henderson2019b}
Matthew Henderson, Ivan Vuli{\'c}, Daniela Gerz, I{\~n}igo Casanueva, Pawe{\l}
  Budzianowski, Sam Coope, Georgios Spithourakis, Tsung-Hsien Wen, Nikola
  Mrk{\v{s}}i{\'c}, and Pei-Hao Su. 2019.
\newblock Training neural response selection for task-oriented dialogue
  systems.
\newblock In \emph{Proceedings of the 57th Annual Meeting of the Association
  for Computational Linguistics}, pages 5392--5404.

\bibitem[{{Hershey} and {Olsen}(2007)}]{KLD-approx-GMM}
J.~R. {Hershey} and P.~A. {Olsen}. 2007.
\newblock \href {https://doi.org/10.1109/ICASSP.2007.366913} {Approximating the
  kullback leibler divergence between gaussian mixture models}.
\newblock In \emph{2007 IEEE International Conference on Acoustics, Speech and
  Signal Processing - ICASSP '07}, volume~4, pages IV--317--IV--320.

\bibitem[{Higgins et~al.(2017)Higgins, adnd Arka~Pal, Burgess, Glorot,
  Botvinick, Mohamed, and Lerchner}]{beta-VAE}
Irina Higgins, Loic~Matthey adnd Arka~Pal, Christopher Burgess, Xavier Glorot,
  Matthew Botvinick, Shakir Mohamed, and Alexander Lerchner. 2017.
\newblock \href {https://openreview.net/forum?id=Sy2fzU9gl} {beta-vae: Learning
  basic visual concepts with a constrained variational framework.}
\newblock In \emph{ICLR}.

\bibitem[{Houlsby et~al.(2019)Houlsby, Giurgiu, Jastrzebski, Morrone,
  De~Laroussilhe, Gesmundo, Attariyan, and Gelly}]{adapters_houlsby19a}
Neil Houlsby, Andrei Giurgiu, Stanislaw Jastrzebski, Bruna Morrone, Quentin
  De~Laroussilhe, Andrea Gesmundo, Mona Attariyan, and Sylvain Gelly. 2019.
\newblock Parameter-efficient transfer learning for {NLP}.
\newblock In \emph{Proceedings of the 36th International Conference on Machine
  Learning}, volume~97 of \emph{Proceedings of Machine Learning Research},
  pages 2790--2799. PMLR.

\bibitem[{Huang et~al.(2019)Huang, Ji, and May}]{huang2019cross}
Lifu Huang, Heng Ji, and Jonathan May. 2019.
\newblock Cross-lingual multi-level adversarial transfer to enhance
  low-resource name tagging.
\newblock In \emph{Proceedings of the 2019 Conference of the North American
  Chapter of the Association for Computational Linguistics: Human Language
  Technologies, Volume 1 (Long and Short Papers)}, pages 3823--3833.

\bibitem[{Jang et~al.(2017)Jang, Gu, and Poole}]{GumbelSoftmax}
Eric Jang, Shixiang Gu, and Ben Poole. 2017.
\newblock \href {https://openreview.net/pdf?id=rkE3y85ee} {Categorical
  reparameterization with gumbel-softmax}.
\newblock In \emph{ICLR}.

\bibitem[{Johnson et~al.(2017)Johnson, Schuster, Le, Krikun, Wu, Chen, Thorat,
  Vi{\'e}gas, Wattenberg, Corrado et~al.}]{johnson2017google}
Melvin Johnson, Mike Schuster, Quoc~V Le, Maxim Krikun, Yonghui Wu, Zhifeng
  Chen, Nikhil Thorat, Fernanda Vi{\'e}gas, Martin Wattenberg, Greg Corrado,
  et~al. 2017.
\newblock Google’s multilingual neural machine translation system: Enabling
  zero-shot translation.
\newblock \emph{Transactions of the Association for Computational Linguistics},
  5:339--351.

\bibitem[{Joulin et~al.(2016)Joulin, Grave, Bojanowski, Douze, J{\'e}gou, and
  Mikolov}]{joulin2016fasttext}
Armand Joulin, Edouard Grave, Piotr Bojanowski, Matthijs Douze, H{\'e}rve
  J{\'e}gou, and Tomas Mikolov. 2016.
\newblock {FastText.zip}: Compressing text classification models.
\newblock \emph{arXiv preprint arXiv:1612.03651}.

\bibitem[{Kannan et~al.(2016)Kannan, Kurach, Ravi, Kaufmann, Tomkins, Miklos,
  Corrado, Luk{\'a}cs, Ganea, Young, and Ramavajjala}]{Kannan2016}
Anjuli Kannan, Karol Kurach, Sujith Ravi, Tobias Kaufmann, Andrew Tomkins,
  Balint Miklos, Gregory~S. Corrado, L{\'a}szl{\'o} Luk{\'a}cs, Marina Ganea,
  Peter Young, and Vivek Ramavajjala. 2016.
\newblock {Smart Reply: Automated Response Suggestion for Email}.
\newblock In \emph{KDD}.

\bibitem[{Kingma and Welling(2014)}]{Kingma2013AutoEncodingVB}
Diederik~P. Kingma and Max Welling. 2014.
\newblock {Auto-Encoding Variational Bayes}.
\newblock \emph{ICLR}.

\bibitem[{Lample and Conneau(2019)}]{lample2019cross}
Guillaume Lample and Alexis Conneau. 2019.
\newblock Cross-lingual language model pretraining.
\newblock \emph{arXiv preprint arXiv:1901.07291}.

\bibitem[{Lee et~al.(2021)Lee, Min, Lee, and Hwang}]{Meta-GMVAE}
Dong~Bok Lee, Dongchan Min, Seanie Lee, and Sung~Ju Hwang. 2021.
\newblock \href {https://openreview.net/forum?id=wS0UFjsNYjn} {Meta-gmvae:
  Mixture of gaussian vae for unsupervised meta-learning}.
\newblock In \emph{ICLR}.

\bibitem[{Lee et~al.(2019)Lee, Tang, and Lin}]{lee2019would}
Jaejun Lee, Raphael Tang, and Jimmy Lin. 2019.
\newblock What would elsa do? freezing layers during transformer fine-tuning.
\newblock \emph{arXiv preprint arXiv:1911.03090}.

\bibitem[{Li et~al.(2020)Li, Gao, Li, Li, Peng, Zhang, and Gao}]{li2020optimus}
Chunyuan Li, Xiang Gao, Yuan Li, Xiujun Li, Baolin Peng, Yizhe Zhang, and
  Jianfeng Gao. 2020.
\newblock \href
  {https://www.microsoft.com/en-us/research/publication/optimus-organizing-sentences-via-pre-trained-modeling-of-a-latent-space/}
  {Optimus: Organizing sentences via pre-trained modeling of a latent space}.

\bibitem[{Lin(2004)}]{lin-2004-rouge}
Chin-Yew Lin. 2004.
\newblock \href {https://www.aclweb.org/anthology/W04-1013} {{ROUGE}: A package
  for automatic evaluation of summaries}.
\newblock In \emph{Text Summarization Branches Out}, pages 74--81, Barcelona,
  Spain. Association for Computational Linguistics.

\bibitem[{{Lin} et~al.(2020){Lin}, {Goyal}, {Girshick}, {He}, and
  {Dollár}}]{focal_loss}
T.~{Lin}, P.~{Goyal}, R.~{Girshick}, K.~{He}, and P.~{Dollár}. 2020.
\newblock \href {https://doi.org/10.1109/TPAMI.2018.2858826} {Focal loss for
  dense object detection}.
\newblock \emph{IEEE Transactions on Pattern Analysis and Machine
  Intelligence}, 42(2):318--327.

\bibitem[{Peters et~al.(2019)Peters, Ruder, and Smith}]{peters2019tune}
Matthew~E Peters, Sebastian Ruder, and Noah~A Smith. 2019.
\newblock To tune or not to tune? adapting pretrained representations to
  diverse tasks.
\newblock In \emph{Proceedings of the 4th Workshop on Representation Learning
  for NLP (RepL4NLP-2019)}, pages 7--14.

\bibitem[{Phang et~al.(2020)Phang, Calixto, Htut, Pruksachatkun, Liu, Vania,
  Kann, and Bowman}]{phang2020english}
Jason Phang, Iacer Calixto, Phu~Mon Htut, Yada Pruksachatkun, Haokun Liu, Clara
  Vania, Katharina Kann, and Samuel Bowman. 2020.
\newblock English intermediate-task training improves zero-shot cross-lingual
  transfer too.
\newblock In \emph{Proceedings of the 1st Conference of the Asia-Pacific
  Chapter of the Association for Computational Linguistics and the 10th
  International Joint Conference on Natural Language Processing}, pages
  557--575.

\bibitem[{Ranasinghe and Zampieri(2020)}]{ranasinghe2020multilingual}
Tharindu Ranasinghe and Marcos Zampieri. 2020.
\newblock Multilingual offensive language identification with cross-lingual
  embeddings.
\newblock In \emph{Proceedings of the 2020 Conference on Empirical Methods in
  Natural Language Processing (EMNLP)}, pages 5838--5844.

\bibitem[{Rao et~al.(2019)Rao, Visin, Rush, Teh, Pascanu, and
  Hadsell}]{CURL-GMVAE}
Dushyant Rao, Francesco Visin, Andrei~A. Rush, Yee~Whye Teh, Razvan Pascanu,
  and Raia Hadsell. 2019.
\newblock \href {https://arxiv.org/abs/1910.14481} {Continual unsupervised
  representation learning}.
\newblock In \emph{NeurIPS}.

\bibitem[{Shang et~al.(2015)Shang, Lu, and Li}]{shang2015neural}
Lifeng Shang, Zhengdong Lu, and Hang Li. 2015.
\newblock Neural responding machine for short-text conversation.
\newblock In \emph{Proceedings of the 53rd Annual Meeting of the Association
  for Computational Linguistics and the 7th International Joint Conference on
  Natural Language Processing (Volume 1: Long Papers)}, pages 1577--1586.

\bibitem[{Shen et~al.(2017)Shen, Su, Li, Li, Niu, Zhao, Aizawa, and
  Long}]{Shen2017ACV}
Xiaoyu Shen, Hui Su, Yanran Li, Wenjie Li, Shuzi Niu, Yang Zhao, Akiko Aizawa,
  and Guoping Long. 2017.
\newblock {A Conditional Variational Framework for Dialog Generation}.
\newblock In \emph{ACL}.

\bibitem[{Shi et~al.(2019)Shi, ad~Brooks~Paige, and Torr}]{Var-MOE}
Yuge Shi, Siddharth~N ad~Brooks~Paige, and Philip Torr. 2019.
\newblock Variational mixture-of-experts autoencoders for multi-modal deep
  generative models.
\newblock In \emph{NeurIPS}.

\bibitem[{Swanson et~al.(2019)Swanson, Yu, Fox, Wohlwend, and
  Lei}]{ProdChatbotRetrieval}
Kyle Swanson, Lili Yu, Christopher Fox, Jeremy Wohlwend, and Tao Lei. 2019.
\newblock \href {https://doi.org/10.18653/v1/W19-4104} {Building a production
  model for retrieval-based chatbots}.
\newblock In \emph{Proceedings of the First Workshop on NLP for Conversational
  AI}, pages 32--41, Florence, Italy. Association for Computational
  Linguistics.

\bibitem[{Tran et~al.(2017)Tran, Haffari, and Zukerman}]{Tran2017AGA}
Quan~Hung Tran, Gholamreza Haffari, and Ingrid Zukerman. 2017.
\newblock {A Generative Attentional Neural Network Model for Dialogue Act
  Classification}.
\newblock In \emph{ACL}.

\bibitem[{Wang et~al.(2020{\natexlab{a}})Wang, Lipton, and
  Tsvetkov}]{wang2020negative}
Zirui Wang, Zachary~C Lipton, and Yulia Tsvetkov. 2020{\natexlab{a}}.
\newblock On negative interference in multilingual language models.
\newblock In \emph{Proceedings of the 2020 Conference on Empirical Methods in
  Natural Language Processing (EMNLP)}, pages 4438--4450.

\bibitem[{Wang et~al.(2020{\natexlab{b}})Wang, Tsvetkov, Firat, and
  Cao}]{gradient-vaccine}
Zirui Wang, Yulia Tsvetkov, Orhan Firat, and Yuan Cao. 2020{\natexlab{b}}.
\newblock Gradient vaccine: Investigating and improving multi-task optimization
  in massively multilingual models.
\newblock In \emph{ICLR}.

\bibitem[{Wei and Deng(2017)}]{VAE-CrossLingual}
Liangchen Wei and Zhi-Hong Deng. 2017.
\newblock \href {https://www.ijcai.org/proceedings/2017/0582.pdf} {A
  variational autoencoding approach for inducing cross-lingual word
  embeddings}.
\newblock In \emph{IJCAI}.

\bibitem[{Wolf et~al.(2020)Wolf, Debut, Sanh, Chaumond, Delangue, Moi, Cistac,
  Rault, Louf, Funtowicz, Davison, Shleifer, von Platen, Ma, Jernite, Plu, Xu,
  Scao, Gugger, Drame, Lhoest, and Rush}]{wolf-etal-2020-transformers}
Thomas Wolf, Lysandre Debut, Victor Sanh, Julien Chaumond, Clement Delangue,
  Anthony Moi, Pierric Cistac, Tim Rault, Rémi Louf, Morgan Funtowicz, Joe
  Davison, Sam Shleifer, Patrick von Platen, Clara Ma, Yacine Jernite, Julien
  Plu, Canwen Xu, Teven~Le Scao, Sylvain Gugger, Mariama Drame, Quentin Lhoest,
  and Alexander~M. Rush. 2020.
\newblock \href {https://www.aclweb.org/anthology/2020.emnlp-demos.6}
  {Transformers: State-of-the-art natural language processing}.
\newblock In \emph{Proceedings of the 2020 Conference on Empirical Methods in
  Natural Language Processing: System Demonstrations}, pages 38--45, Online.
  Association for Computational Linguistics.

\bibitem[{Yang et~al.(2019)Yang, Cheung, Li, and Fang}]{VAE-GMM-Graphs}
Linxiao Yang, Ngai-Man Cheung, Jiaying Li, and Jun Fang. 2019.
\newblock \href
  {https://openaccess.thecvf.com/content_ICCV_2019/papers/Yang_Deep_Clustering_by_Gaussian_Mixture_Variational_Autoencoders_With_Graph_Embedding_ICCV_2019_paper.pdf}
  {Deep clustering by gaussian mixture variational autoencoders with graph
  embedding}.
\newblock In \emph{ICCV}.

\bibitem[{Yang et~al.(2020)Yang, Cer, Ahmad, Guo, Law, Constant, Abrego, Yuan,
  Tar, Sung, Strope, and Kurzweil}]{MLUniEnc}
Yinfei Yang, Daniel Cer, Amin Ahmad, Mandy Guo, Jax Law, Noah Constant,
  Gustavo~Hernandez Abrego, Steve Yuan, Chris Tar, Yun-Hsuan Sung, Brian
  Strope, and Ray Kurzweil. 2020.
\newblock \href {https://www.aclweb.org/anthology/2020.acl-demos.12.pdf}
  {Multilingual universal sentence encoder for semantic retrieval}.
\newblock In \emph{ACL}.

\bibitem[{Ying et~al.(2021)Ying, Bajaj, Deb, Yang, Wang, Lin, Shokouhi, Song,
  Yang, and Jiang}]{QiYingSRNAACL2021}
Qianlan Ying, Payal Bajaj, Budhaditya Deb, Yu~Yang, Wei Wang, Bojia Lin, Milad
  Shokouhi, Xia Song, Yang Yang, and Daxin Jiang. 2021.
\newblock Language scaling for universal suggested replies model.
\newblock In \emph{NAACL-HLT, Industrial Track}.

\bibitem[{Yu et~al.(2020)Yu, Wu, Zeng, Tao, Deng, and Jiang}]{crossing-VAE-QA}
Wenhao Yu, Lingfei Wu, Qingkai Zeng, Shu Tao, Yu~Deng, and Meng Jiang. 2020.
\newblock \href {https://doi.org/10.18653/v1/2020.acl-main.498} {Crossing
  variational autoencoders for answer retrieval}.
\newblock In \emph{Proceedings of the 58th Annual Meeting of the Association
  for Computational Linguistics}, pages 5635--5641, Online. Association for
  Computational Linguistics.

\bibitem[{Zhang et~al.(2016)Zhang, Xiong, Su, Duan, and Zhang}]{VAE-MT}
Biao Zhang, Deyi Xiong, Jinsong Su, Hong Duan, and Min Zhang. 2016.
\newblock \href {https://doi.org/10.18653/v1/D16-1050} {Variational neural
  machine translation}.
\newblock In \emph{Proceedings of the 2016 Conference on Empirical Methods in
  Natural Language Processing}, pages 521--530.

\bibitem[{Zhang et~al.(2021)Zhang, Wang, Deb, Zheng, Shokouhi, and
  Awadallah}]{zhang-etal-2021-dataset}
Mozhi Zhang, Wei Wang, Budhaditya Deb, Guoqing Zheng, Milad Shokouhi, and
  Ahmed~Hassan Awadallah. 2021.
\newblock \href {https://doi.org/10.18653/v1/2021.acl-long.97} {A dataset and
  baselines for multilingual reply suggestion}.
\newblock In \emph{Proceedings of the 59th Annual Meeting of the Association
  for Computational Linguistics and the 11th International Joint Conference on
  Natural Language Processing (Volume 1: Long Papers)}, pages 1207--1220,
  Online. Association for Computational Linguistics.

\bibitem[{Zhao et~al.(2017)Zhao, Zhao, and Esk{\'e}nazi}]{Zhao2017LearningDD}
Tiancheng Zhao, Ran Zhao, and Maxine Esk{\'e}nazi. 2017.
\newblock {Learning Discourse-level Diversity for Neural Dialog Models using
  Conditional Variational Autoencoders}.
\newblock In \emph{ACL}.

\bibitem[{Zhou et~al.(2016)Zhou, Dong, Wu, Zhao, Yu, Tian, Liu, and
  Yan}]{zhou-etal-2016-multi}
Xiangyang Zhou, Daxiang Dong, Hua Wu, Shiqi Zhao, Dianhai Yu, Hao Tian, Xuan
  Liu, and Rui Yan. 2016.
\newblock \href {https://doi.org/10.18653/v1/D16-1036} {Multi-view response
  selection for human-computer conversation}.
\newblock In \emph{Proceedings of the 2016 Conference on Empirical Methods in
  Natural Language Processing}, pages 372--381, Austin, Texas. Association for
  Computational Linguistics.

\bibitem[{Zhou et~al.(2018)Zhou, Li, Dong, Liu, Chen, Zhao, Yu, and
  Wu}]{zhou-etal-2018-multi}
Xiangyang Zhou, Lu~Li, Daxiang Dong, Yi~Liu, Ying Chen, Wayne~Xin Zhao, Dianhai
  Yu, and Hua Wu. 2018.
\newblock \href {https://doi.org/10.18653/v1/P18-1103} {Multi-turn response
  selection for chatbots with deep attention matching network}.
\newblock In \emph{Proceedings of the 56th Annual Meeting of the Association
  for Computational Linguistics (Volume 1: Long Papers)}, pages 1118--1127,
  Melbourne, Australia. Association for Computational Linguistics.

\end{thebibliography}
    \begin{appendix}

    \begin{figure*}
    	\centering
    	\includegraphics[scale=0.57,trim={0 80 0 80},clip]{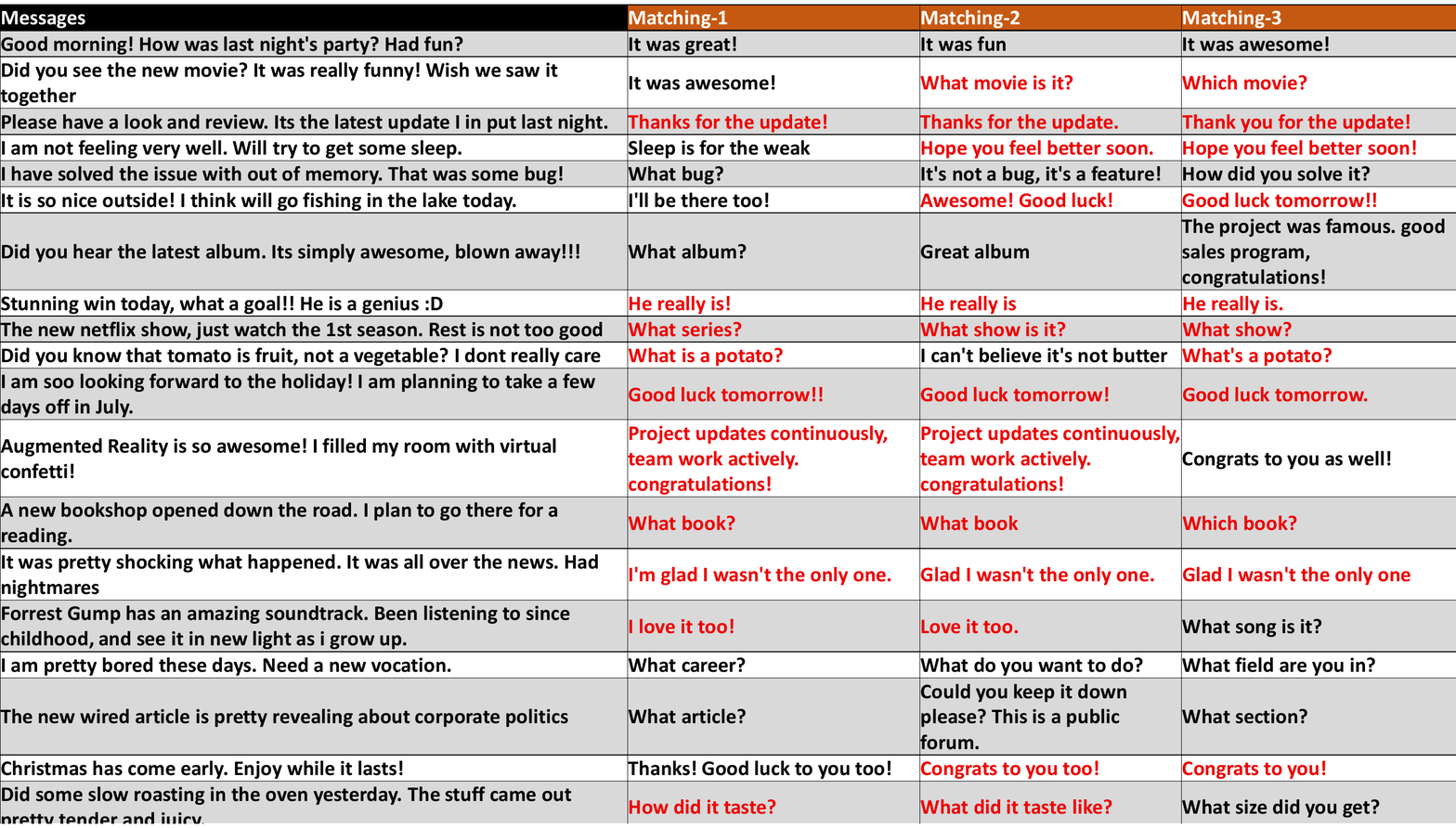}
    	\caption{Some samples of English message predicted with English replies using the Matching Model. The replies marked in red shows the duplicate responses.} 
    	\label{tab:Samples_EN_Matching}
    \end{figure*}
    
    \begin{figure*}
    	\centering
    	\includegraphics[scale=0.57,trim={0 80 0 80},clip]{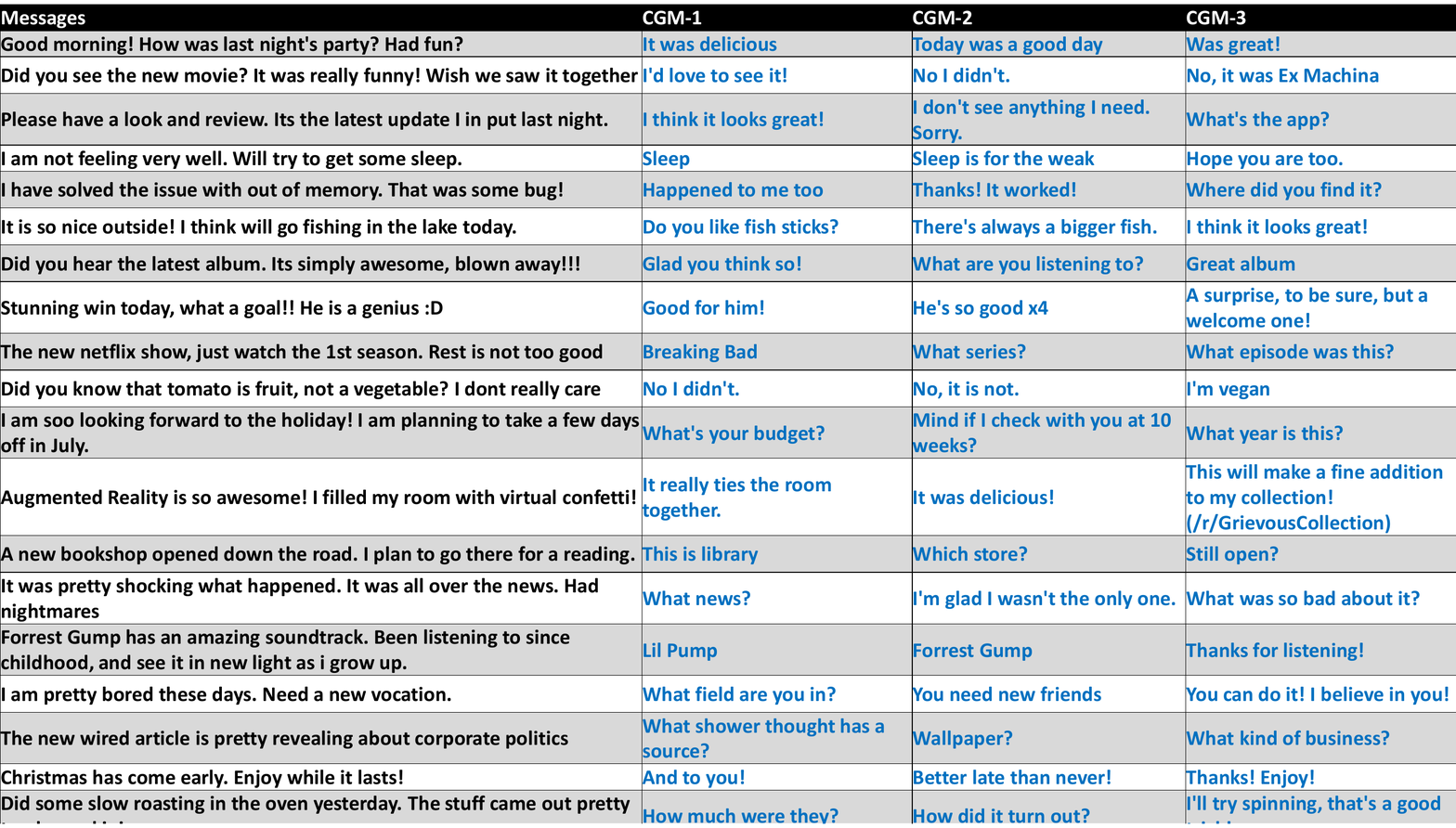}
    	\caption{Some samples of English message predicted with English replies using the CGM Model.}
    	\label{tab:Samples_EN_CGM}
    \end{figure*}
    
        \begin{figure*}
    	\centering
    	\includegraphics[scale=0.57,trim={0 80 0 80},clip]{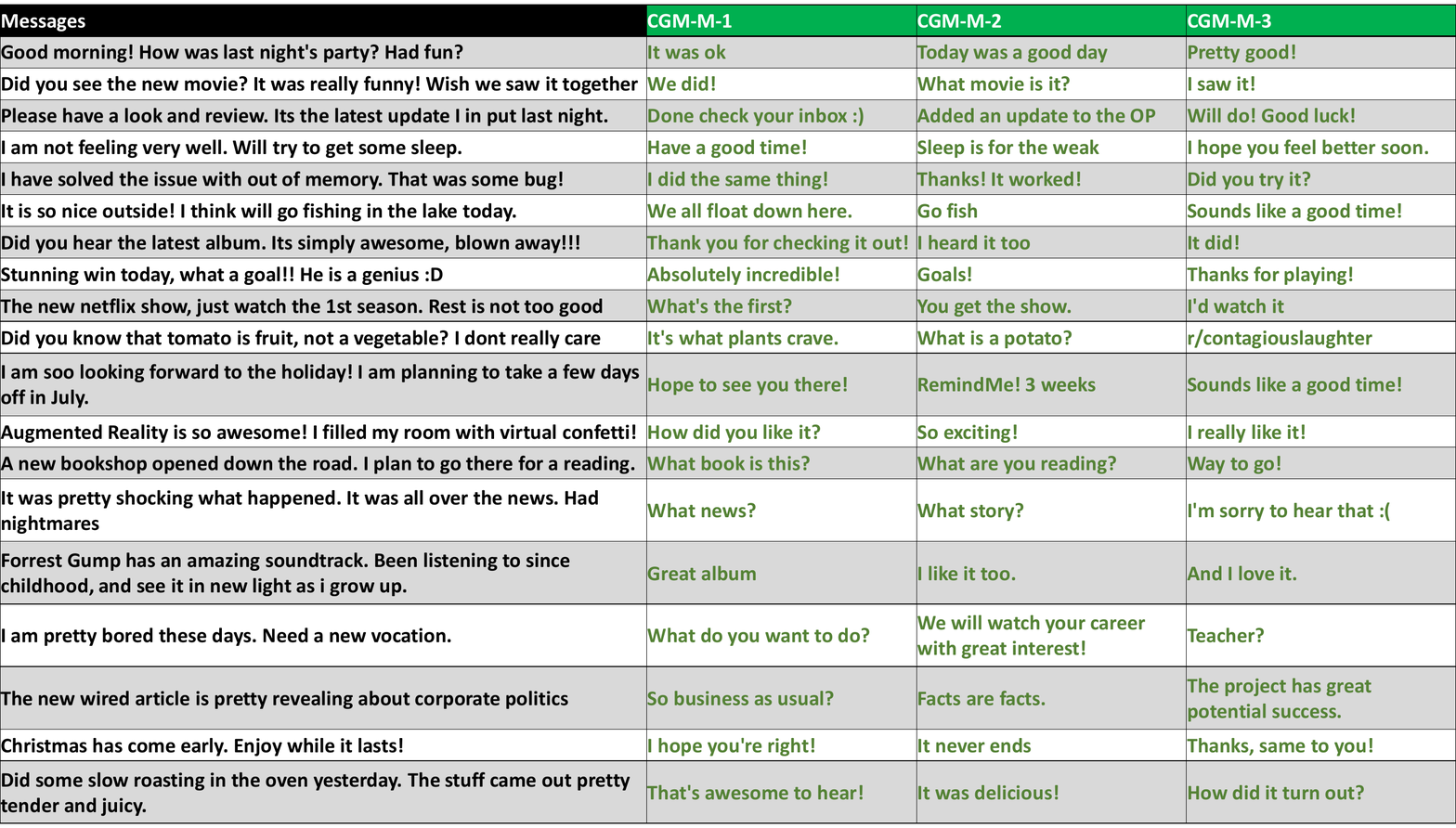}
    	\caption{Some samples of English message predicted with English replies using the  CGM-M Model.}
    	\label{tab:Samples_EN_CGM_M}
    \end{figure*}
    
\section{Text Samples from Model Predictions}


\subsection{Relevance and Diversity}
We created sample messages in EN manually, and predict the responses from different models: Matching in Figure \ref{tab:Samples_EN_Matching}, CGM in Figure \ref{tab:Samples_EN_CGM} and CGM-M in Figure \ref{tab:Samples_EN_CGM_M}. 

We see that in terms of relevance while it is hard to notice the differences on such a small sample, overall the predictions from the Matching model are less relevant than CGM.  However, we can clearly distinguish the diversity of responses: predictions from Matching have a high level of duplicates where some of the responses differ by just a punctuation. While this can be easily de-duplicated using simple heuristics, the results show that inherently the Matching model ranks very similar responses at the top. The CGM models in contrast, show a lot of diversity in responses without reducing the relevance of the responses. 

We also see that some of the responses are quite specific and not relevant, with some responses being rude or mildly inappropriate. It shows the issues with using responses from the Reddit dataset without careful curation (the MRS dataset does clean up for inappropriate responses but cannot completely eliminate them without human curation). Production systems usually have human curated response sets which can tackle these issues. 

    \begin{figure*}
    	\centering
    	\includegraphics[scale=0.57,trim={0 80 0 80},clip]{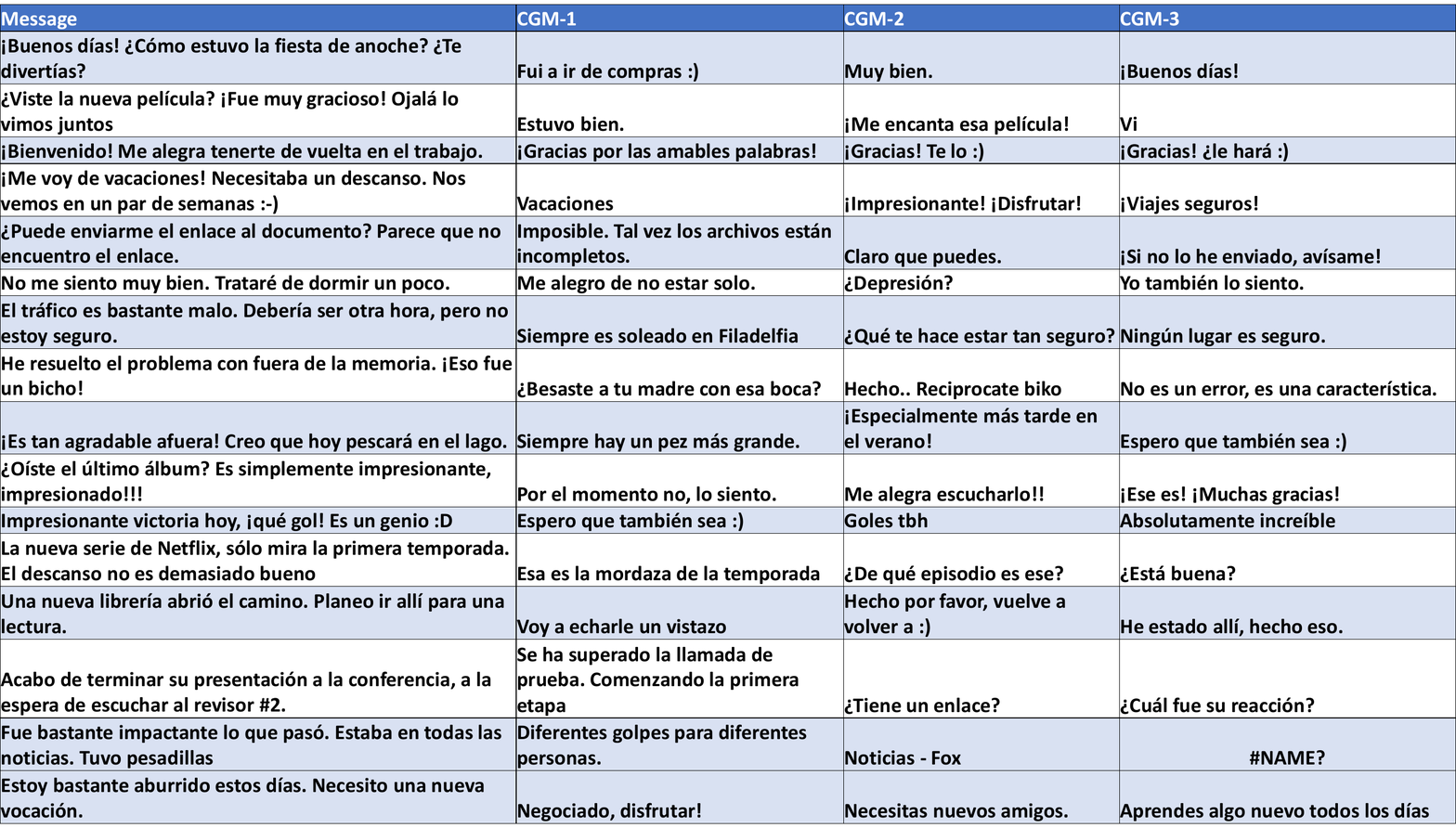}
    	\caption{Some samples of Spanish messages and predicted with Spanish replies using the CGM-M Model.}
    	\label{tab:Samples_ES_to_ES_CGM_M}
    \end{figure*}
    
    \begin{figure*}
    	\centering
    	\includegraphics[scale=0.57,trim={0 80 0 80},clip]{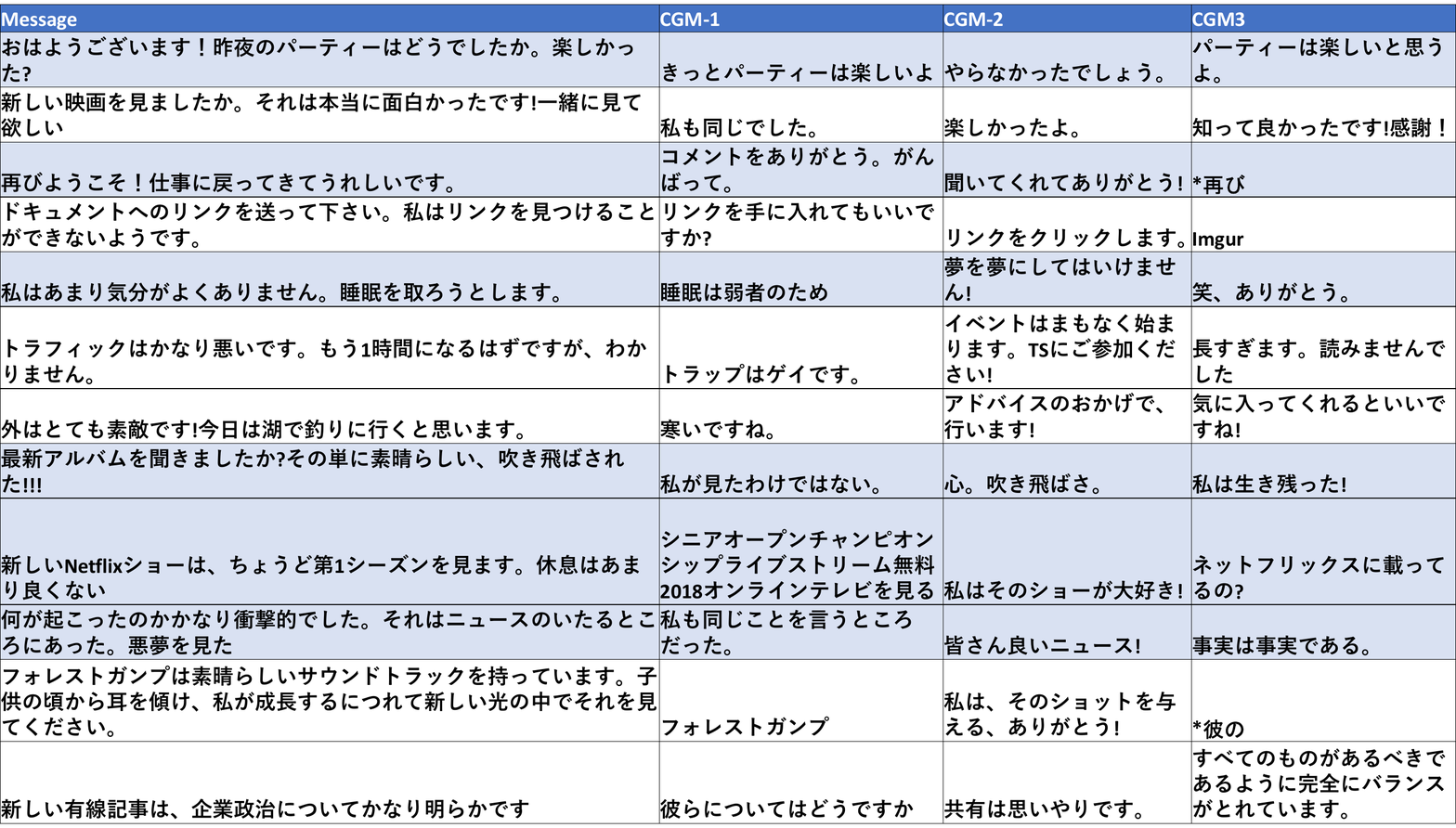}
    	\caption{Some samples of Japanese messages and predicted with Japanese replies using the CGM-M Model.}
    	\label{tab:Samples_JA_to_JA_CGM_M}
    \end{figure*}
    
    \begin{figure*}
    	\centering
    	\includegraphics[scale=0.57,trim={0 80 0 80},clip]{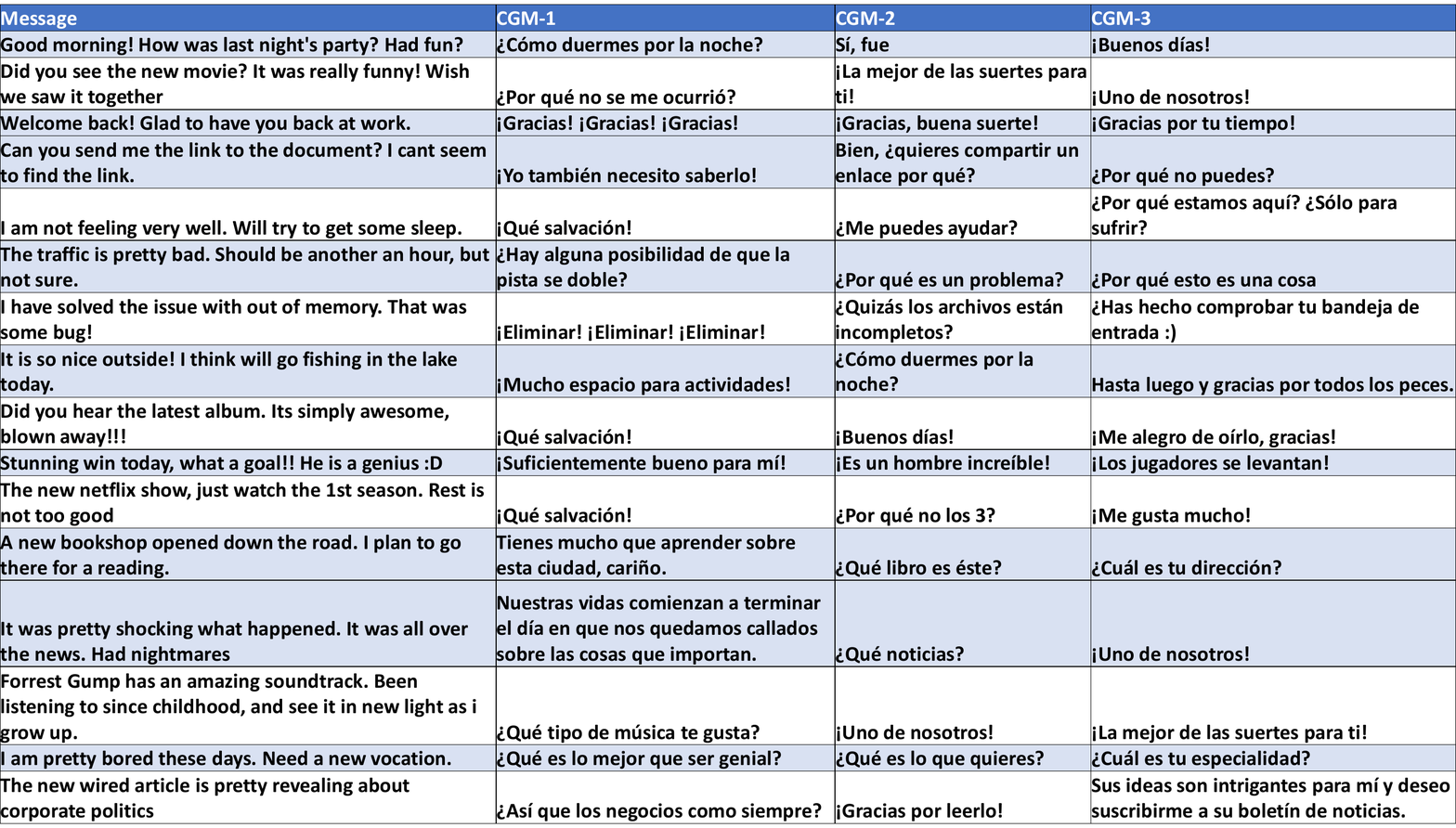}
    	\caption{Some samples of English messages and predicted with Spanish replies using the CGM-M Model. While the quality is not as good as when the input message is in Spanish, the general close match of intents of the message and responses illustrates the cross lingual ability of of the model.}
    	\label{tab:Samples_EN_to_ES_CGM_M}
    \end{figure*}
    
     \begin{figure*}
    	\centering
    	\includegraphics[scale=0.57,trim={0 80 0 80},clip]{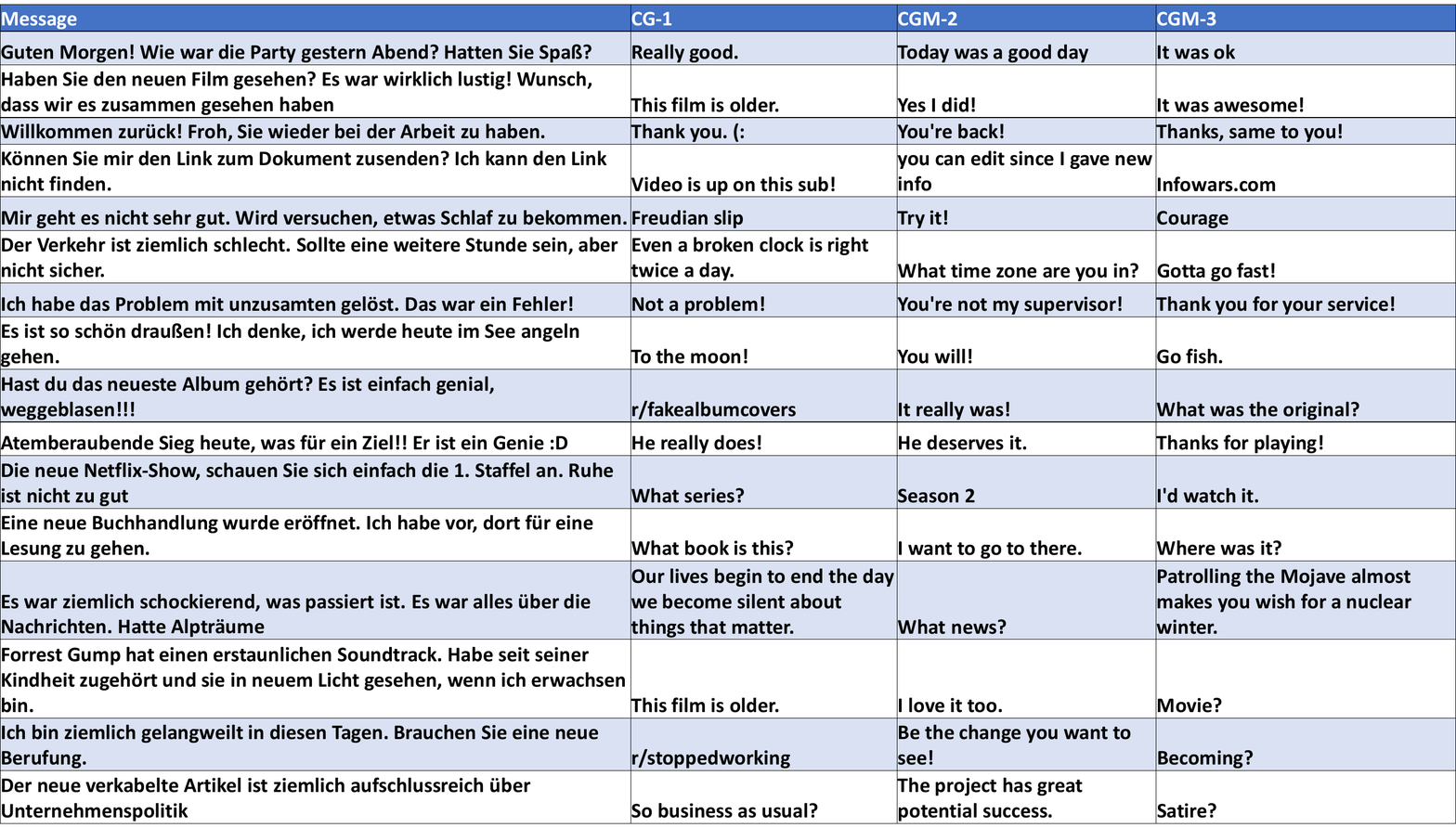}
    	\caption{Some samples of German messages and predicted with English replies using the CGM-M Model. While the quality is not as good as when the input message is in German, the general close match of intents of the message and responses illustrates the cross lingual ability of of the model.}
    	\label{tab:Samples_DE_to_EN_CGM_M}
    \end{figure*}
    
    \begin{figure*}
    	\centering
    	\includegraphics[scale=0.57,trim={0 80 0 80},clip]{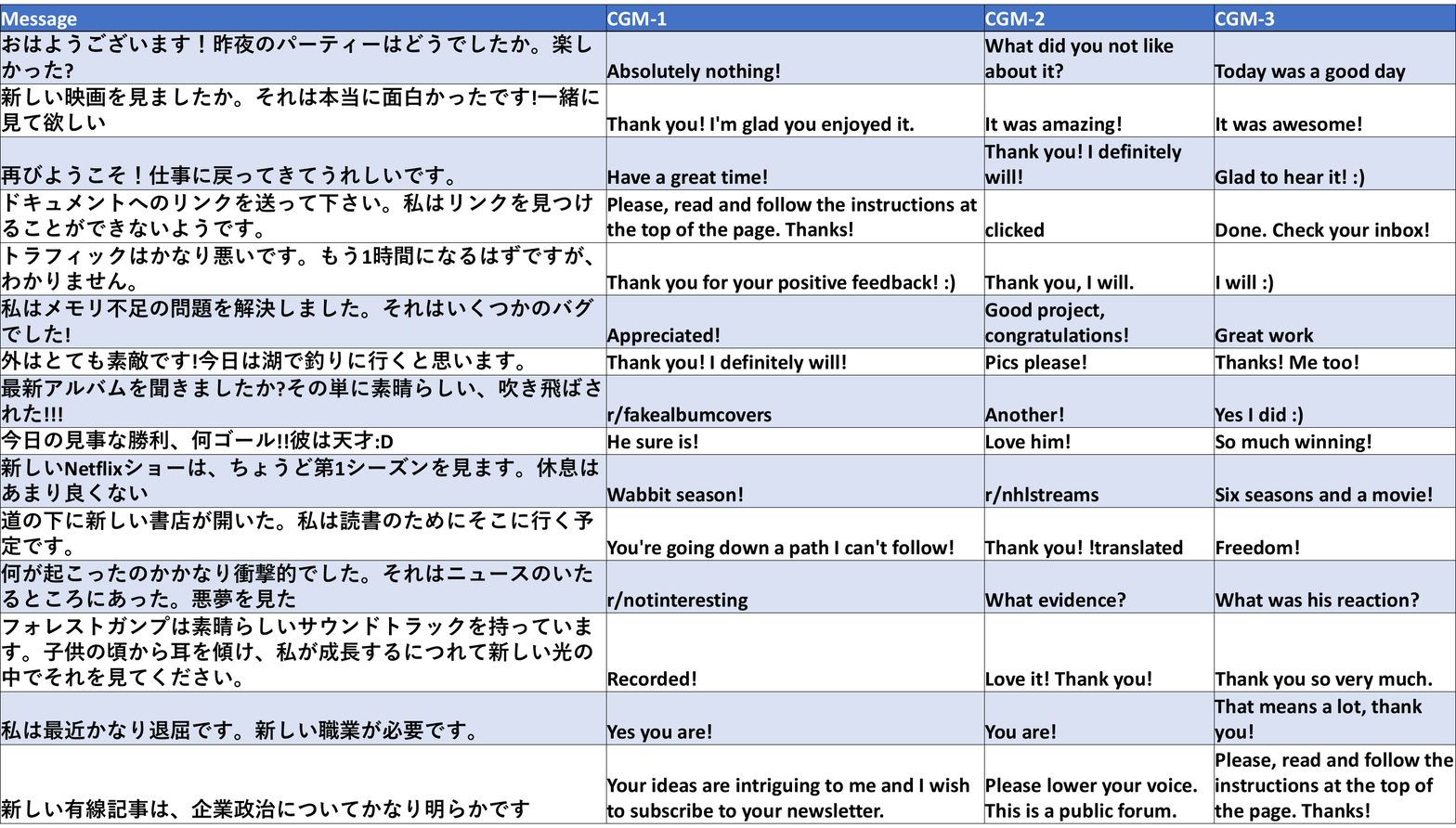}
    	\caption{Some samples of Japanese messages and predicted with English replies using the CGM-M Model. The quality here is definitely poorer that German to English, perhaps since EN and JA are not as closely related. However we still get the general close match of intents of the message and responses.} 
    	\label{tab:Samples_JA_to_EN_CGM_M}
    \end{figure*}

\subsection{Multi-lingual Behavior}
Next we look at the multilingual ability of CGM. We translate the same set of messages used for EN for predicting responses, so as to have better comparative understanding of the quality different languages.

We present the predictions for \textbf{ES} in Fig. \ref{tab:Samples_ES_to_ES_CGM_M} and \textbf{JA} in Fig. \ref{tab:Samples_JA_to_JA_CGM_M}.  We see that the responses are relevant and diverse in these languages and thus CGM performs adequately in languages other than EN.

\subsection{Cross-lingual Behavior}
Finally we investigate the cross lingual nature of the CGM model, in order to understand if the multi-lingual models share representations and learnings across languages. 

In Figure \ref{tab:Samples_EN_to_ES_CGM_M} we use EN messages and force the model to predict responses from the ES set. Surprisingly such a system is able to select relevant results in the target language. While the quality here is not as good, but it is interesting to see that such cross lingual prediction works quite well. 

In Figure \ref{tab:Samples_DE_to_EN_CGM_M} we use messages in German and predict with English responses. Again the results are quite acceptable. This may be expected as English and German are closely related languages. To see slightly different pairs of languages, we look at Japanese messages with predictions in English in Figure \ref{tab:Samples_JA_to_EN_CGM_M}. Here the quality is actually much worse, but we still see some match with the overall intent of the messages showing good cross lingual representation even for weakly related languages.

    \end{appendix}
\end{document}